\documentclass{article}

\usepackage[utf8]{inputenc} 
\usepackage[T1]{fontenc}    
\usepackage[colorlinks=true,citecolor=blue]{hyperref}       
\usepackage{url}            
\usepackage{booktabs}       
\usepackage{amsfonts}       
\usepackage{nicefrac}       
\usepackage{microtype}      
\usepackage{centernot}
\usepackage[normalem]{ulem}

\usepackage[left=1.25in, top=1in, bottom=1in, right=1.25in]{geometry}
\usepackage[utf8]{inputenc} 
\usepackage[T1]{fontenc}    
\usepackage{url}            
\usepackage{booktabs}       
\usepackage{amsfonts}       
\usepackage{nicefrac}       
\usepackage{microtype}      

\usepackage{natbib}

\usepackage{float,subcaption}
\usepackage{graphicx}
\usepackage[dvipsnames,usenames]{xcolor}
\usepackage[export]{adjustbox}
\usepackage[inline]{enumitem}

\usepackage{tikz}
\usetikzlibrary{positioning}

\usepackage{amsmath,amssymb,amsthm}
\usepackage[mathscr]{eucal}
\usepackage{stmaryrd}
\usepackage{accents}
\usepackage{upgreek}

\usepackage[most]{tcolorbox}

\newtcolorbox{mybox}[1][]{colback=yellow!5, colframe=black, fonttitle=\bfseries, boxsep=1pt, left=2pt,right=2pt,top=-5pt,bottom=1pt}

\newcommand{\arxiv}[2]{#2}

\usepackage{booktabs,colortbl,multirow}

\usepackage{algorithm,algorithmic}

\graphicspath{{figs//}}

\allowdisplaybreaks

\newcommand{\best}[1]{\cellcolor{gray!25}{#1}}

\newcommand{\newedit}[1]{#1}

\newcommand{\defEq}{\stackrel{.}{=}}

\newcommand{\indicator}[1]{\llbracket #1 \rrbracket}

\newcommand{\tick}{$\checkmark$}
\newcommand{\cross}{$\times$}

\newcommand{\argmax}{{\operatorname{argmax }}}

\newcommand{\argmaxUnique}[2]{{\operatorname{argmax }\nolimits_{#1}}\, #2}

\renewcommand{\Pr}{\mathbb{P}}
\newcommand{\E}[2]{{\mathbb{E}}_{#1}\left[ #2 \right]}

\newcommand{\X}{{x}}
\newcommand{\Y}{{y}}

\newcommand{\XCal}{\mathscr{X}}
\newcommand{\YCal}{\mathscr{Y}}

\newcommand{\Real}{\mathbb{R}}

\newcommand{\expp}{{\sc Exp }}

\newtheorem{lemma}{Lemma}

\newtheorem{theorem}[lemma]{Theorem}

\theoremstyle{definition}

\newtcbtheorem[number within=section]%
{remark_tcb} 
{Remark} 
{%
fonttitle=\bfseries\upshape,fontupper=\normalsize,
colframe=green!10!black,colback=green!2!white,
colbacktitle=green!20!white,coltitle=blue!75!black,
boxsep=1pt,left=2pt,right=2pt,top=1pt,bottom=1pt,
frame hidden,boxrule=1pt,
theorem style=plain} 
{rem} 

\newtcbtheorem[number within=section]%
{assumption_tcb} 
{Assumption} 
{%
fonttitle=\bfseries\upshape,fontupper=\normalsize,
colframe=blue!5!black,colback=blue!2!white,
colbacktitle=blue!10!white,coltitle=blue!75!black,
boxsep=1pt,left=2pt,right=2pt,top=1pt,bottom=1pt,
frame hidden,boxrule=1pt,
theorem style=plain} 
{asm} 

\title{Long-Tail Learning via Logit Adjustment}

\author{Aditya Krishna Menon \and Sadeep Jayasumana \and Ankit Singh Rawat \and Himanshu Jain \qquad Andreas Veit \qquad Sanjiv Kumar \\
Google Research, New York \\
\texttt{\{adityakmenon,sadeep,ankitsrawat,himj,aveit,sanjivk\}@google.com}}

\begin{document}

\maketitle

\begin{abstract}
Real-world classification problems typically exhibit an \emph{imbalanced} or \emph{long-tailed} label distribution,
wherein many labels are associated with only a few samples.
This poses a challenge for generalisation on such labels,
and also 
makes na\"{i}ve learning biased towards dominant labels.
In this paper, we present two 
simple modifications of standard softmax cross-entropy training 
to cope with these challenges.
Our techniques revisit the classic idea of \emph{logit adjustment} based on the label frequencies,
either applied post-hoc to a trained model,
or enforced in the loss during training.
Such adjustment encourages a large \emph{relative margin} between logits of rare versus dominant labels.
These techniques unify and generalise several recent proposals in the literature,
while possessing firmer 
statistical grounding
and empirical performance. A reference implementation of our methods is available at:

\vspace{0.1cm}
\noindent
{\small\url{https://github.com/google-research/google-research/tree/master/logit\_adjustment}.}

\end{abstract}

\section{Introduction}
Real-world classification problems typically exhibit a
\emph{long-tailed} label distribution,
wherein most labels are associated with only a few samples~\citep{VanHorn:2017,Buda:2017,Liu:2019}.
Owing to this paucity of samples,
generalisation on such labels is challenging;
moreover,
na\"{i}ve learning on such data is susceptible to an undesirable bias towards dominant labels.
This problem has been widely studied in the literature on
learning under \emph{class imbalance}~\citep{CardieHo97,Chawla:2002,QiaoLi09,HeGa09,Wallace:2011}%
\arxiv{.}{
and
\emph{the related problem of cost-sensitive learning}~\citep{Elkan01,Zadrozny:2001,Masnadi-Shirazi:2010,Dmochowski+10}.}

Recently,
long-tail learning has received renewed interest in the context of neural networks.
Two active strands of work involve
post-hoc normalisation of the classification weights~\citep{Zhang:2019,Kim:2019,Kang:2020,Ye:2020},
and
modification of the underlying loss to account for varying class penalties~\citep{Zhang:2017,Cui:2019,Cao:2019,Tan:2020}.
Each of these strands is intuitive, and has proven empirically successful.
However, they are not without limitation:
e.g., weight normalisation crucially relies on the weight norms being smaller for rare classes; however, this assumption is sensitive to the choice of optimiser (see~\S\ref{sec:weight-norm-adam}).
On the other hand,
loss modification sacrifices the \emph{consistency} that underpins the softmax cross-entropy (see~\S\ref{sec:unified-margin}).
Consequently,
existing techniques may result in suboptimal solutions
even in simple settings (\S\ref{sec:synth-expt}).

In this paper, 
we present two 
simple modifications of softmax cross-entropy training 
that unify several 
recent
proposals,
and overcome their limitations.
\arxiv{Our techniques involve
\emph{logit adjustment} based on label frequencies,}
{Our techniques revisit the classic idea of
\emph{logit adjustment} based on label frequencies~\citep{Provost:2000,Zhou:2006,Collell:2016},}
applied either 
post-hoc on a trained model,
or
as a modification of the training loss.
Conceptually, logit adjustment 
encourages a large \emph{relative margin} 
between a pair of rare and dominant labels.
This has
a firm statistical grounding:
unlike recent techniques, it is \emph{consistent} for 
minimising 
the 
\emph{balanced error}~(cf.~\eqref{eqn:ber}),
a common 
metric
in long-tail settings
which averages the per-class errors.
This 
grounding
translates into
strong
empirical performance on real-world datasets. 

In summary, our contributions are:
\begin{enumerate*}[label=(\roman*),itemsep=0pt,topsep=0pt,leftmargin=16pt]

    \item 
    we present two 
    realisations of logit adjustment for long-tail learning,
    applied either post-hoc (\S\ref{sec:post-hoc-logit}) or during training (\S\ref{sec:unified-margin})

    \item
    we establish that logit adjustment overcomes limitations in
    recent proposals (see Table~\ref{tbl:summary}),
    and in particular is \emph{Fisher consistent}
    for
    minimising 
    the 
    \emph{balanced error}~(cf.~\eqref{eqn:ber});

    \item we confirm the efficacy of 
    the 
    proposed
    techniques 
    on 
    real-world datasets (\S\ref{sec:experiments}).
\end{enumerate*}
\arxiv{}{In the course of our analysis, we also present a general version of the softmax cross-entropy with a \emph{pairwise label margin}~\eqref{eqn:unified-margin-loss}, which offers flexibility in controlling the relative contribution of labels to the overall loss.}

\begin{table}[!t]
    \centering
    
    \renewcommand{\arraystretch}{1.25}
    \resizebox{0.99\linewidth}{!}{
    \begin{tabular}{@{}lllp{1.5in}@{}}
        \toprule
        \textbf{Method} & \textbf{Procedure} & \textbf{Consistent?} & \textbf{Reference} \\
        \toprule
        Weight normalisation & Post-hoc weight scaling 
        & \cross & \citep{Kang:2020} \\
        Adaptive margin & Softmax with \newedit{rare} +ve upweighting 
        & \cross & \citep{Cao:2019} \\
        Equalised margin & Softmax with \newedit{rare} -ve downweighting 
        & \cross & \citep{Tan:2020} \\
        \midrule
        Logit-adjusted threshold & Post-hoc logit translation 
        & \tick & This paper (cf.~\eqref{eqn:logit-adjustment}) \\
        Logit-adjusted loss & Softmax with logit translation
        & \tick & This paper (cf.~\eqref{eqn:logit-adjusted-loss}) \\
        \bottomrule
    \end{tabular}
    }
    \caption{
    Comparison of approaches to long-tail learning. 
    Weight normalisation
    re-scales the classification weights;
    by contrast, we \emph{add} per-label offsets to the logits.
    {
    Margin approaches 
    \newedit{uniformly}
    increase the margin 
    between a {rare} positive and 
    \newedit{all negatives}%
    ~\citep{Cao:2019}, or decrease the margin between 
    \newedit{all positives} 
    and a {rare} negative~\citep{Tan:2020}
    to prevent suppression of rare labels' gradients. 
    \newedit{By contrast, we increase the margin
    between a \emph{rare} positive and a \emph{dominant} negative.}
    }
    }
    \label{tbl:summary}
    \arxiv{\vspace{-1.5\baselineskip}}{}
\end{table}

\section{Problem setup and related work}
\label{sec:background}

%
\label{sec:long-tail-bg}

Consider a multiclass classification problem with instances $\XCal$ and labels $\YCal = [ L ] \defEq \{ 1, 2, \ldots, L \}$.
Given a sample $S = \{ ( x_n, y_n ) \}_{n = 1}^N \sim \Pr^{N}$, for unknown distribution $\Pr$ over $\XCal \times \YCal$,
our goal is to learn a scorer $f \colon \XCal \to \Real^L$ that minimises the 
misclassification error 
$ 
{\Pr}_{\X, \Y}\Big( y \notin \argmaxUnique{y' \in \YCal}{ f_{y'}( x ) } \Big). $
Typically, 
one minimises
a surrogate loss
$\ell \colon \YCal \times \Real^L \to \Real$,
such as the softmax cross-entropy, 
\begin{equation}
    \label{eqn:softmax-xent}
    \ell( y, f( x ) ) = \log \Big[ \sum\nolimits_{y' \in [L]} e^{f_{y'}( x )} \Big] - f_y( x ) = \log\Big[ 1 + \sum\nolimits_{y' \neq y} e^{f_{y'}( x ) - f_{y}( x )} \Big].    
\end{equation}
For $p_y( x ) \propto {e^{f_{y}( x )}}$, we may view $p(x) \defEq [p_{1}(x),\ldots, p_L(x)] \in \Delta_{|\YCal|}$
as an estimate of
$\Pr( \Y \mid \X )$.

The setting of \emph{learning under class imbalance} 
or \emph{long-tail learning}
is where the distribution $\Pr(y)$ is highly skewed, so that 
many (rare or ``tail'') labels have a very low probability of occurrence.
Here, the misclassification error is not a suitable measure of performance:
a trivial predictor which
classifies every instance 
to the majority label will attain a low misclassification error.
To cope with this, a natural alternative is the balanced error~\citep{Chan:98,Brodersen:2010,Menon:2013},
which averages each of the per-class error rates:
\begin{equation}
    \label{eqn:ber}
    \mathrm{BER}( f ) \defEq 
\frac{1}{L} \sum\nolimits_{y \in [L]} {\Pr}_{\X|\Y}\Big( y \notin \argmaxUnique{y' \in \YCal}{ f_{y'}( \X ) } \Big).
\end{equation}
This can be seen as implicitly 
using
a \emph{balanced} class-probability function
    $
    \Pr^{\mathrm{bal}}( y \mid x ) \propto \frac{1}{L} \cdot \Pr( x \mid y ),$
as opposed to the native $\Pr( y \mid x ) \propto \Pr( y ) \cdot \Pr( x \mid y )$ that is employed in the misclassification error.

%
\label{sec:approaches}

Broadly, extant approaches to coping with class imbalance 
\arxiv{(see also Table~\ref{tbl:literature} in the Appendix)}{(see also Table~\ref{tbl:literature})}
modify:
\begin{enumerate}[label=(\roman*),itemsep=0pt,topsep=0pt,leftmargin=16pt]
    \item the \emph{inputs} to a model,
    for example
    by over- or under-sampling~\citep{KubatMa97,Chawla:2002,Wallace:2011,Mikolov:2013,Mahajan:2018,Yin:2018}
    
    \item the \emph{outputs} of a model, for example by post-hoc correction of the decision threshold
    ~\citep{Fawcett:1996,Collell:2016}
    or weights~\citep{Kim:2019,Kang:2020}
    
    \item the \emph{internals} of a model, for example by modifying the loss function~\citep{Xie:1989,Morik:1999,Cui:2019,Zhang:2017,Cao:2019,Tan:2020}
\end{enumerate}

\arxiv{}{%
\begin{table}[!ht]
    \centering
    \begin{tabular}{@{}llp{2.75in}@{}}
        \toprule
        {\bf Family} & {\bf Method} & {\bf Reference} \\
        \toprule
        Post-hoc correction   & Modify threshold & \citep{Fawcett:1996,Provost:2000,Maloof03,King:2001,Collell:2016} \\
        & Normalise weights & \citep{Zhang:2019,Kim:2019,Kang:2020} \\
        \midrule
        Data modification & Under-sampling & \citep{KubatMa97,Wallace:2011} \\
        & Over-sampling & \citep{Chawla:2002} \\
        & Feature transfer & \citep{Yin:2018} \\
        \midrule
        Loss weighting & Loss balancing & \citep{Xie:1989,Morik:1999,Menon:2013} \\
        & Volume weighting & \citep{Cui:2019} \\ 
        & Average top-$k$ loss & \citep{Fan:2017} \\
        & Domain adaptation & \citep{Jamal:2020} \\
        \midrule
        Margin modification & Cost-sensitive SVM & \citep{Masnadi-Shirazi:2010,Iranmehr:2019} \\
        & Range loss & \citep{Zhang:2017} \\
        & Label-aware margin & \citep{Cao:2019}
        \\
        & Equalised negatives & \citep{Tan:2020} \\
         \bottomrule
    \end{tabular}
    \caption{Summary of different approaches to learning under class imbalance.}
    \label{tbl:literature}
\end{table}
}

One may easily combine
approaches from the first stream with those from the latter two.
Consequently,
we focus on the latter two in this work,
and describe some representative recent examples from each.

\textbf{Post-hoc weight normalisation}.
Suppose $f_y( x ) = w_y^{\top} \Phi( x )$ for classification weights $w_y \in \Real^D$ and representations $\Phi \colon \XCal \to \Real^D$,
as learned by a neural network.
(We may add per-label bias terms to $f_y$ by adding a constant feature to $\Phi$.)
A fruitful avenue of exploration 
involves decoupling of representation and classifier learning~\citep{Zhang:2019}.
Concretely,
we
first learn $\{w_y, \Phi \}$ via standard training on the long-tailed training sample $S$,
and then predict for $x \in \XCal$
\begin{equation}
    \label{eqn:weight-normalisation}
    \argmaxUnique{y \in [L]}{ {w_y^{\top} \Phi( x )} / { \nu_y^{\tau} } } = \argmaxUnique{y \in [L]}{ {f_{y}( x )} / { \nu_y^{\tau} } },
\end{equation}
for $\tau > 0$,
where 
$\nu_y = \Pr( y )$ in~\citet{Kim:2019,Ye:2020}
and
$\nu_y = \| w_y \|_2$ in~\citet{Kang:2020}.
Further to the above,
one may also enforce $\| w_y \|_2 = 1$ during training~\citep{Kim:2019}. 
Intuitively, 
either choice of $\nu_y$
upweights the contribution of rare labels
through \emph{weight normalisation}.
The choice $\nu_y = \| w_y \|_2$ is motivated by the observations that $\| w_y \|_2$ tends to correlate with $\Pr( y )$.

%
\begin{figure}[!t]
    \centering
    \subcaptionbox{CIFAR-10-LT.}{\includegraphics[scale=0.325]{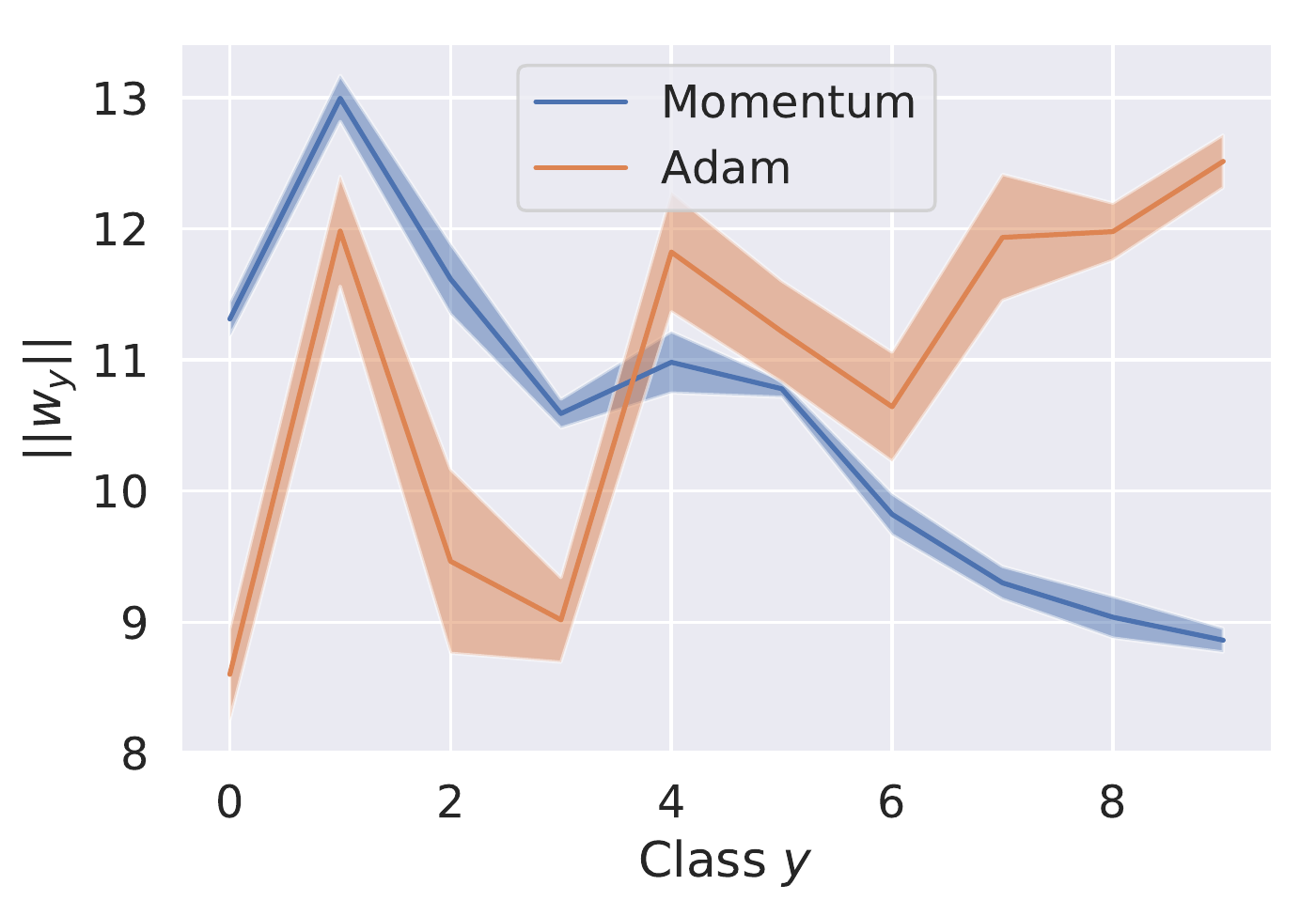}}%
    \qquad%
    \subcaptionbox{CIFAR-100-LT.}{\includegraphics[scale=0.325]{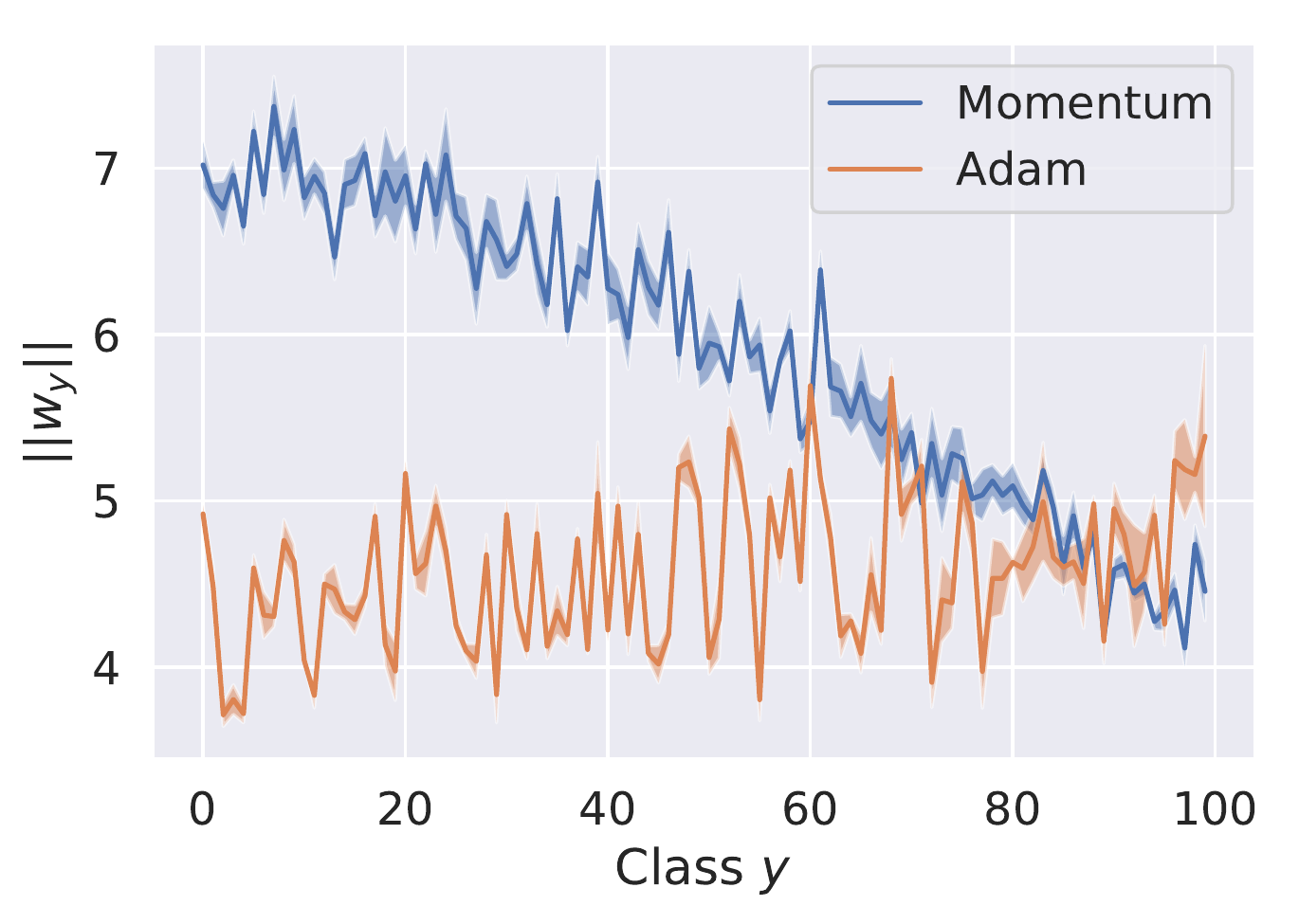}}
    
    \caption{Mean and standard deviation 
    over $5$ runs
    of 
    per-class weight norms 
    for a ResNet-32
    under momentum and Adam optimisers.
    We use long-tailed (``LT'') versions of CIFAR-10 and CIFAR-100,
    and sort classes in descending order of frequency;
    the first class is {100} times more likely to appear than the last class.
    Both optimisers yield solutions with comparable balanced error. 
    However,
    the weight norms have 
    incompatible trends:
    under momentum, the norms are strongly correlated with class frequency,
    while with Adam, the norms are
    \emph{anti-correlated} or
    \emph{independent} of the class frequency.
    Consequently,
    weight normalisation under Adam is ineffective for combatting class imbalance. 
    }
    \label{fig:weight_norm_cifar10}
    \arxiv{\vspace{-\baselineskip}}{}
\end{figure}

\textbf{Loss modification}. %
A classic means of coping with class imbalance is to 
\emph{balance} the loss,
wherein $\ell( y, f( x ) )$ is weighted by $\Pr( y )^{-1}$~\citep{Xie:1989,Morik:1999}:
for example, 
\begin{equation}
    \label{eqn:balanced-loss}
    \ell( y, f( x ) ) = \frac{1}{\Pr( y )} \cdot \log\Big[ 1 + \sum\nolimits_{y' \neq y} e^{f_{y'}( x ) - f_{y}( x )} \Big].
\end{equation}
While intuitive,
balancing has minimal effect in separable settings:
solutions that achieve zero training loss 
will necessarily remain optimal even under weighting~\citep{Byrd:2019}.
Intuitively, one would like instead to shift the separator closer to a dominant class.
\citet{Li:2002,Wu:2008,Masnadi-Shirazi:2010,Iranmehr:2019,Gottlieb:2020} thus proposed to 
add \emph{per-class margins} into the hinge loss.
\citep{Cao:2019} proposed to add a per-class margin into the softmax cross-entropy:
\begin{equation}
    \label{eqn:cao}
    \ell( y, f(x) ) = \log\Big[ 1 + \sum\nolimits_{y' \neq y} e^{\delta_y} \cdot e^{f_{y'}(x) - f_{y}(x)} \Big],
\end{equation}
where $\delta_y \propto \Pr( y )^{-1/4}$.
This 
{upweights rare ``positive'' labels $y$, which} enforces a larger margin between a {rare positive} $y$ and any ``negative'' $y' \neq y$.
Separately,~\citet{Tan:2020} proposed
\begin{equation}
    \label{eqn:equalised}
    \ell( y, f(x) ) = \log\Big[ 1 + \sum\nolimits_{y' \neq y} e^{\delta_{y'}} \cdot e^{f_{y'}(x) - f_{y}(x)} \Big],
\end{equation}
where 
$\delta_{y'} \leq 0$ is an non-decreasing transform of $\Pr( y' )$.
\newedit{The motivation is that\newedit{, in the original softmax cross-entropy without $\{\delta_{y'}\}$, a rare label often receives a strong \emph{inhibitory} gradient signal as it disproportionately appear as a negative for dominant labels.}}
See also~\citet{Liu:2016,Liu:2017,Wang:2018,Khan:2019} for similar weighting of negatives in the softmax.

\textbf{Limitations of existing approaches}.
Each of the above methods are intuitive,
and have shown strong empirical performance.
However, a closer analysis identifies some subtle limitations.

\emph{Limitations of weight normalisation}.
Post-hoc weight normalisation 
with 
$\nu_y = \| w_y \|_2$ per~\citet{Kang:2020} is 
motivated by the observation that
the weight norm $\| w_y \|_2$ tends to correlate with $\Pr( y )$.
However, 
we now show
this assumption
is highly dependent on the choice of optimiser.




%
\label{sec:weight-norm-adam}
We consider long-tailed versions of CIFAR-10 and CIFAR-100,
wherein the first class is 100 times more likely to appear than the last class.
(See~\S\ref{sec:experiments-real} for more details on these datasets.)
We optimise a ResNet-32 using both SGD with momentum and Adam optimisers.
Figure~\ref{fig:weight_norm_cifar10} 
confirms that under SGD, $\| w_y \|_2$ and the class priors $\Pr( y )$ are correlated.
However,
with Adam, 
the norms are 
either 
\emph{anti}-correlated 
or 
\emph{independent} of the class priors.
This marked difference 
may be understood in light of recent study of the {implicit bias} of optimisers~\citep{Soudry:2018};
cf. Appendix~\ref{sec:normalisation-margins}.
One may hope to side-step this by simply using $\nu_y = \Pr( y )$;
unfortunately,
even this choice has limitations (see~\S\ref{sec:weight-norm-critique}).

\emph{Limitations of loss modification}.
Enforcing
a per-label margin per~\eqref{eqn:cao} and~\eqref{eqn:equalised} 
is intuitive,
as it allows for shifting the decision boundary away from rare classes.
However, when doing so, it is important to ensure 
\emph{Fisher consistency}~\citep{Lin:2004}
(or \emph{classification calibration}~\citep{Bartlett+06})
of the resulting loss for the balanced error.
That is, 
the minimiser of the expected loss 
(equally, the empirical risk in the infinite sample limit)
should result in a minimal balanced error.
Unfortunately, both~\eqref{eqn:cao} and~\eqref{eqn:equalised} are \emph{not} consistent in this sense,
even for binary problems%
;
see~\S\ref{sec:unified-margin},~\S\ref{sec:synth-expt} for
details.




%
\section{Logit adjustment for long-tail learning: a statistical view}
\label{sec:statistical}
The above suggests there is scope for improving performance on long-tail problems,
both in terms of post-hoc correction and loss modification.
We now show how a statistical perspective on the problem suggests simple procedures of each type,
both of which overcome the limitations discussed above.

Recall that our goal is to minimise the balanced error~\eqref{eqn:ber}.
A natural question is:
what 
is the \emph{best possible} or \emph{Bayes-optimal} scorer
for this problem,
i.e.,
$ f^* \in \operatorname{argmin}_{f \colon \XCal \to \Real^L}{ \mathrm{BER}( f ) }. $
Evidently,
such an $f^*$ must depend on the (unknown) underlying distribution $\Pr( x, y )$.
Indeed,
we have%
~\citep{Menon:2013},~\citep[Theorem 1]{Collell:2016}
\begin{equation}
    \label{eqn:bayes-ber}
    \argmaxUnique{y \in [L]}{ f^*_y( x ) } = 
    \argmaxUnique{y \in [L]}{ \Pr^{\mathrm{bal}}( y \mid x ) } =
    \argmaxUnique{y \in [L]}{ \Pr( x \mid y ) },
\end{equation}
where $\Pr^{\mathrm{bal}}$ is the balanced class-probability as per~\S\ref{sec:long-tail-bg}. 
In words, 
the Bayes-optimal prediction is the label under which the given instance $x \in \XCal$ is most likely.
Consequently, for fixed class-conditionals $\Pr( x \mid y )$,
varying the \emph{class priors} $\Pr( y )$ arbitrarily will not affect the optimal scorers.
This is intuitively desirable: the balanced error is agnostic to the level of imbalance in the label distribution.

To further probe~\eqref{eqn:bayes-ber},
suppose the underlying class-probabilities
$\Pr( y \mid x ) \propto \exp( s^*_y( x ) )$, for (unknown) scorer 
$s^* \colon \XCal \to \Real^L$.
Since by definition
$\Pr^{\mathrm{bal}}( y \mid x ) \propto {\Pr( y \mid x )} / {\Pr( y )}$,~\eqref{eqn:bayes-ber} becomes
\begin{equation}
    \label{eqn:bayes-logit-adjustment}
    \argmaxUnique{y \in [L]}{ \Pr^{\mathrm{bal}}( y \mid x ) } =
    \argmaxUnique{y \in [L]}{ {\exp( s^*_y( x ) )} / {\Pr( y )} } =
    \argmaxUnique{y \in [L]}{ s^*_y( x ) - \ln \Pr( y ) },
\end{equation}
i.e.,
we translate the (unknown) distributional scores or logits based on the class priors.
This simple fact immediately suggests two means of optimising for the balanced error:
\begin{enumerate}[label=(\roman*),itemsep=0pt,topsep=0pt,leftmargin=16pt]
    \item train a model 
    to estimate the standard $\Pr( y \mid x )$
    (\newedit{e.g., by minimising the standard softmax-cross entropy on the long-tailed data}),
    and then explicitly modify its logits post-hoc as per~\eqref{eqn:bayes-logit-adjustment}
    
    \item train a model to estimate the balanced $\Pr^{\mathrm{bal}}( y \mid x )$,
    whose logits are implicitly modified as per~\eqref{eqn:bayes-logit-adjustment}
\end{enumerate}
Such \emph{logit adjustment} techniques 
\arxiv{}{--- which have been a classic approach to class-imbalance~\citep{Provost:2000} ---}
neatly align with the post-hoc and loss modification streams discussed in~\S\ref{sec:background}.
However, 
unlike most previous techniques from these streams,
logit adjustment is endowed with a clear statistical grounding:
by construction, 
the optimal solution under such adjustment coincides with the Bayes-optimal solution~\eqref{eqn:bayes-ber} for the balanced error,
i.e.,
it is \emph{Fisher consistent} for minimising the balanced error.
We shall demonstrate this translates into superior empirical performance (\S\ref{sec:experiments}).
Note also that logit adjustment
may be easily extended to cover performance measures beyond the balanced error,
e.g.,
with distinct costs for errors on dominant and rare classes;
we leave a detailed study and contrast to existing cost-sensitive approaches~\citep{Iranmehr:2019,Gottlieb:2020} to future work.

We now study each of the techniques (i) and (ii) in turn.

%
\section{Post-hoc logit adjustment}
We now detail to perform post-hoc logit adjustment on a classifier trained on long-tailed data.
We further show this 
bears similarity to recent weight normalisation schemes,
but has a 
subtle advantage.

\subsection{The post-hoc logit adjustment procedure}
\label{sec:logit-adjustment}

Given a sample $S \sim \Pr^N$ of long-tailed data,
suppose we learn a neural network
with logits
$f_y( x ) = w_y^{\top} \Phi( x )$.
Given these,
one typically predicts the label 
$\operatorname{argmax}_{y \in [L]}{ f_{y}( x ) }$.
When trained with the softmax cross-entropy, one may view $p_y( x ) \propto \exp( f_y( x ) )$ as an approximation of the underlying $\Pr( y \mid x )$,
and so this equivalently predicts the label with highest estimated class-probability.

In \emph{post-hoc logit adjustment},
we propose to instead predict, for suitable $\tau > 0$: 
\begin{equation}
    \label{eqn:logit-adjustment}
    \argmaxUnique{y \in [L]}{ {\exp( w_y^{\top} \Phi( x ) )} / { \pi_y^{\tau} } } =
    \argmaxUnique{y \in [L]}{ f_{y}( x ) - {\tau \cdot \log \pi_{y}} },
\end{equation}
where
$\pi \in \Delta_{\YCal}$ are estimates of the class priors $\Pr( y )$, e.g.,
the empirical class frequencies on the training sample $S$.
Effectively,~\eqref{eqn:logit-adjustment} adds a label-dependent offset to each of the logits.
When $\tau = 1$, this can be seen as 
applying~\eqref{eqn:bayes-logit-adjustment}
with
a plugin estimate of $\Pr( y \mid x )$,
i.e.,
${p}_y( x ) \propto \exp( w_y^{\top} \Phi( x ) )$.
When $\tau \neq 1$, 
this can be seen as 
applying~\eqref{eqn:bayes-logit-adjustment}
to \emph{temperature scaled}
estimates
    $\bar{p}_y( x ) \propto \exp( \tau^{-1} \cdot w_y^{\top} \Phi( x ) ).$
To unpack this,
recall that~\eqref{eqn:bayes-logit-adjustment}
justifies post-hoc logit thresholding
given access to the true probabilities $\Pr( y \mid x )$.
In principle, the outputs of a sufficiently high-capacity neural network aim to mimic these probabilities.
In practice, 
these estimates
are often uncalibrated~\citep{Guo:2017}.
One may thus need to first calibrate the probabilities before applying logit adjustment.
Temperature scaling 
is 
one means of doing so, and is often used in the context of distillation~\citep{Hinton:2015}.

One may 
treat $\tau$ as a tuning parameter
to be chosen based on some measure of holdout calibration,
\newedit{
e.g., the expected calibration error~\citep{Murphy:1987,Guo:2017}, 
probabilistic sharpness~\citep{Gneiting:2007b,Kuleshov:2018},
or a proper scoring rule such as the log-loss or squared error~\citep{Gneiting:2007}.
One may alternately fix $\tau = 1$
and aim to learn inherently calibrated probabilities,
e.g.,
via label smoothing~\citep{Szegedy:2016,Muller:2019}.
}

\subsection{Comparison to existing post-hoc techniques}
\label{sec:weight-norm-critique}

Post-hoc logit adjustment with $\tau = 1$ is not a new idea in the class imbalance literature.
Indeed, this is a standard technique when creating stratified samples~\citep{King:2001},
and 
when training
binary classifiers
~\citep{Fawcett:1996,Provost:2000,Maloof03}.
In multiclass settings,
this has been explored in%
~\citet{Zhou:2006,Collell:2016}.
However, $\tau \neq 1$ is important in practical usage of neural networks, owing to their lack of calibration.
Further,
we now explicate that
post-hoc logit adjustment
has an important advantage over 
recent post-hoc weight normalisation techniques.

Recall that weight normalisation involves learning a 
scorer
$f_y( x ) = w_y^{\top} \Phi( x )$,
and then post-hoc normalising the weights via 
${w_y} / {\nu_y^{\tau}}$ for $\tau > 0$.
We demonstrated in~\S\ref{sec:background} that using $\nu_y = \| w_y \|_2$ may be ineffective when using adaptive optimisers.
However, even with $\nu_y = \pi_y$,
there is a subtle contrast to post-hoc logit adjustment:
while the former performs a \emph{multiplicative}
update to the logits,
the latter performs an \emph{additive} update.
The two techniques may thus yield different orderings over labels, since
\[ \frac{w_1^\mathrm{T} \Phi( x )}{\pi_1} < \frac{w_2^\mathrm{T} \Phi( x )}{\pi_2} < \cdots < \frac{w_L^\mathrm{T} \Phi( x )}{\pi_L} \substack{\centernot\implies\\\centernot\impliedby}
\frac{e^{w_1^\mathrm{T} \Phi( x )}}{\pi_1} < \frac{e^{w_2^\mathrm{T} \Phi( x )}}{\pi_2} < \cdots < \frac{e^{w_L^\mathrm{T} \Phi( x )}}{\pi_L}.
\]

%
\label{sec:post-hoc-logit}

Weight normalisation
is thus \emph{not} consistent for minimising the balanced error,
unlike logit adjustment.
Indeed, if a rare label $y$
has \emph{negative} score $w_{y}^{\top} \Phi( x ) < 0$, {and there is another label with positive score},
then it is \emph{impossible} for the weight normalisation to give $y$ the highest score.
By contrast, 
under logit adjustment, 
$w_y^\mathrm{T} \Phi( x ) - \ln \pi_y$ will be lower for dominant classes,
regardless of the original sign.

\section{The logit adjusted softmax cross-entropy}
We now show how to directly bake logit adjustment into the softmax cross-entropy.
We show that this approach has an intuitive relation to existing loss modification techniques. 

\subsection{The logit adjusted loss}

From~\S\ref{sec:statistical},
the second approach to optimising for the balanced error is to directly model $\Pr_{\mathrm{bal}}( y \mid x ) \propto {\Pr( y \mid x )} / {\Pr( y )}$.
To do so, 
consider the following \emph{logit adjusted softmax cross-entropy loss}
for $\tau > 0$: 
\begin{equation}
    \label{eqn:logit-adjusted-loss}
    \resizebox{0.9\linewidth}{!}{
    $\displaystyle
    {\ell}( y, f( x ) ) 
    = -\log \frac{e^{f_{y}( x ) + \tau \cdot \log \pi_{y}}}{\sum_{y' \in [L]} e^{f_{y'}( x ) + \tau \cdot \log \pi_{y'}}}
    = \log\bigg[ 1 + \sum\nolimits_{y' \neq y} \left( \frac{\pi_{y'}}{\pi_{y}} \right)^{{\tau}} \cdot e^{(f_{y'}( x ) - f_{y}( x ))} \bigg].
    $}
\end{equation}
Given a scorer that minimises the above, we now predict $\argmax_{y \in [L]} f_y(x)$ as usual.

Compared to the standard softmax cross-entropy~\eqref{eqn:softmax-xent},
the above applies a \emph{label-dependent offset} to each logit.
Compared to~\eqref{eqn:logit-adjustment},
we \emph{directly} enforce the class prior offset while learning the logits,
rather than doing this post-hoc.
\arxiv{From a consistency perspective, these approaches are indistinguishable.
However, from a \emph{generalisation} perspective, we expect~\eqref{eqn:logit-adjusted-loss} to be more useful:
by enforcing a bias during training,
we prevent convergence to degenerate solutions.}
{The two approaches have a deeper connection: observe that~\eqref{eqn:logit-adjusted-loss} is equivalent to using a scorer of the form $g_y( x ) = f_y( x ) + \tau \cdot \log \pi_y$.
We thus have $\argmax_{y \in [L]} f_y(x) = \argmax_{y \in [L]} g_y(x) - \tau \cdot \log \pi_y$.
Consequently, one can equivalently view learning with this loss as learning a standard scorer $g( x )$, and post-hoc adjusting its logits to make a prediction.}
\arxiv{}{For convex objectives, we thus do not expect any difference between the solutions of the two approaches.
For non-convex objectives, as encountered in neural networks, the bias endowed by adding $\tau \cdot \log \pi_y$ to the logits is however likely to result in a different local minima.
}

For more insight into the loss, 
consider the following \emph{pairwise margin loss}
    \begin{align}
        \label{eqn:unified-margin-loss}
        \ell( y, f( x ) ) &= \alpha_y \cdot \log\Big[ 1 + \sum\nolimits_{y' \neq y} e^{\Delta_{y y'}} \cdot e^{(f_{y'}( x ) - f_{y}( x ))} \Big],
    \end{align}
for label weights $\alpha_y > 0$,
and
\emph{pairwise label margins}
$\Delta_{y y'}$
representing the
desired gap between scores for $y$ and $y'$.
For $\tau = 1$, our
logit adjusted loss~\eqref{eqn:logit-adjusted-loss}
corresponds to~\eqref{eqn:unified-margin-loss} with
$\alpha_y = 1$ and
$\Delta_{y y'} = \log\left( \frac{\pi_{y'}}{\pi_y} \right) 
$.
\newedit{This demands a larger margin between \emph{rare} positive ($\pi_y \sim 0$) and \emph{dominant} negative ($\pi_{y'} \sim 1$) labels,
so that scores for dominant classes do not overwhelm those for rare ones.}

\subsection{Comparison to existing loss modification techniques}
\label{sec:unified-margin}

A cursory inspection of~\eqref{eqn:cao},~\eqref{eqn:equalised}
reveals a striking similarity to 
our logit adjusted softmax cross-entropy~\eqref{eqn:logit-adjusted-loss}.
The balanced loss~\eqref{eqn:balanced-loss} also bears similarity, except that the weighting is performed \emph{outside} the logarithm.
Each of these losses
are
special cases of the 
pairwise margin loss~\eqref{eqn:unified-margin-loss}
enforcing \emph{uniform} margins that only consider the positive or negative label,
unlike our approach.

For example, 
$\alpha_y = \frac{1}{\pi_{y}}$ and $\Delta_{y y'} = 0$ yields the balanced loss~\eqref{eqn:balanced-loss}.
This does not explicitly enforce a margin between the labels,
which is undesirable for separable problems~\citep{Byrd:2019}.
When $\alpha_y = 1$, 
the choice $\Delta_{y y'} = \pi_{y}^{-1/4}$ yields~\eqref{eqn:cao}.
Finally,
$\Delta_{y y'} = \log F( \pi_{y'} )$
yields~\eqref{eqn:equalised},
where 
$F \colon [0, 1] \to ( 0, 1 ]$ is some non-decreasing function,
e.g., $F( z ) = z^\tau$ for $\tau > 0$.
These losses 
thus either consider the frequency of the positive $y$
or negative $y'$, but not \emph{both} simultaneously.

The above choices of $\alpha$ and $\Delta$ are all intuitively plausible.
However,~\S\ref{sec:statistical} indicates that our loss in~\eqref{eqn:logit-adjusted-loss} has a firm statistical grounding:
it ensures Fisher consistency for the balanced error.

%
\begin{theorem}
\label{thm:multiclass-unified-consistent}
For any $\delta \in \mathbb{R}_+^L$,
the pairwise loss in~\eqref{eqn:unified-margin-loss} is Fisher consistent with
weights and margins
    \begin{align*}
        \alpha_y &= {\delta_y}/{\pi_y} \qquad \Delta_{y y'} = \log \big({\delta_{y'}}/{\delta_{y}}\big).
    \end{align*}
\end{theorem}

Observe that when $\delta_y = \pi_y$, we immediately deduce that the logit-adjusted loss of~\eqref{eqn:logit-adjusted-loss} is consistent.
Similarly, $\delta_y = 1$ recovers the classic result that the balanced loss is consistent.
While the above is only a sufficient condition, it turns out that in the binary case,
one may neatly encapsulate a necessary and sufficient condition for consistency that rules out other choices;
see Appendix~\ref{sec:binary-consistency}.
This suggests that existing proposals may thus underperform
\arxiv{}{with respect to the balanced error}
in certain settings, as verified empirically in~\S\ref{sec:synth-expt}.


%
\subsection{Discussion and extensions}
\label{sec:further-discussion}

\newedit{
One may be tempted to combine the logit adjusted loss in~\eqref{eqn:logit-adjusted-loss} with the post-hoc adjustment of~\eqref{eqn:logit-adjustment}.
However, following~\S\ref{sec:statistical}, such an approach would not be statistically coherent.
Indeed, minimising a logit adjusted loss encourages the model to estimate the balanced class-probabilities $\Pr^{\mathrm{bal}}( y \mid x )$.
Applying post-hoc adjustment 
will distort these probabilities,
and is thus expected to be \emph{harmful}.

More broadly, however, there is value in combining logit adjustment with other techniques.
For example,
Theorem~\ref{thm:multiclass-unified-consistent} implies that it is sensible to combine logit adjustment with loss weighting;
e.g., one may pick 
$\Delta_{yy'} = {\tau} \cdot \log \left( {{\pi_{y'}}/{\pi_y}} \right)$,
and
$\alpha_y = \pi_y^{\tau - 1}$.
This is similar to~\citet{Cao:2019}, who found benefits in combining weighting with their loss.
One may also
generalise the formulation in Theorem~\ref{thm:multiclass-unified-consistent},
and employ $\Delta_{yy'} = \tau_1 \cdot \log \pi_y - \tau_2 \cdot \log \pi_{y'}$,
where $\tau_{1}, \tau_{2}$ are constants.
This
interpolates
between the logit adjusted loss ($\tau_1 = \tau_2$) and a version of the equalised margin loss ($\tau_1 = 0$).
}

\arxiv{}{\citet[Theorem 2]{Cao:2019} provides a rigorous generalisation bound for the adaptive margin loss
under the assumption of separable training data with binary labels.
The inconsistency of the loss with respect to the balanced error concerns the more general scenario of non-separable multiclass data data, which may occur, e.g., owing to label noise or limitation in model capacity.
We shall subsequently demonstrate that encouraging consistency can lead to gains in practical settings. 
We shall further see that \emph{combining} the $\Delta$ implicit in this loss with our proposed $\Delta$ can lead to further gains,
indicating a potentially complementary nature of the losses.}

Interestingly, for $\tau = -1$, a similar loss to~\eqref{eqn:logit-adjusted-loss} has been considered in the context of \emph{negative sampling} for scalability~\citep{Yi:2019}:
here, one samples a small subset of negatives based on the class priors $\pi$, and applies logit correction {to obtain an unbiased estimate of the unsampled loss function based on all the negatives~\citep{Bengio:2008}}.
Losses of the general form~\eqref{eqn:unified-margin-loss} have also been explored for structured prediction~\citep{Zhang:2005,Pletscher:2010,Hazan:2010}.

\section{Experimental results}
\label{sec:experiments}
We now present experiments confirming our main claims:
\begin{enumerate*}[label=(\roman*),itemsep=0pt,topsep=0pt,leftmargin=16pt]
    \item on simple binary problems, existing weight normalisation and loss modification techniques 
    may not converge to the optimal solution (\S\ref{sec:synth-expt});
    
    \item on real-world datasets, 
    our post-hoc logit adjustment 
    outperforms weight normalisation,
    \newedit{
    and
    one can obtain further gains 
    via our logit adjusted softmax cross-entropy
    (\S\ref{sec:experiments-real}).}
\end{enumerate*}

\subsection{Results on synthetic dataset}
\label{sec:synth-expt}

We 
consider 
a binary classification task,
wherein samples from class $y \in \{ \pm 1 \}$ are drawn from a 2D Gaussian
with 
isotropic covariance
and means $\mu_{y} = y \cdot ( +1, +1 )$.
We introduce class imbalance by setting $\Pr( y = +1 ) = 5\%$.
The Bayes-optimal classifier 
for the balanced error is (see Appendix~\ref{app:gaussian_ber})
\begin{equation}
    \label{eqn:bayes-gaussian}
    f^*( x ) = +1 \iff \Pr( x \mid y = +1 ) > \Pr( x \mid y = -1 ) \iff (\mu_1 - \mu_{-1})^{\top} x > 0,
\end{equation}
i.e., it is a linear separator passing through the origin.
We compare this separator against those found by 
several 
margin
losses
based on~\eqref{eqn:unified-margin-loss}:
standard ERM ($\Delta_{y y'} = 0$),
the adaptive loss~\citep{Cao:2019} ($\Delta_{y y'} = \pi_y^{-1/4}$),
an instantiation of the equalised loss~\citep{Tan:2020} ($\Delta_{y y'} = \log \pi_{y'}$),
and our logit adjusted loss ($\Delta_{y y'} = \log \frac{\pi_{y'}}{\pi_{y}}$).
For each loss, we train an affine classifier on a sample of $10,000$ instances,
and evaluate the balanced error on a test set of $10,000$ samples
over $100$ independent trials.

\begin{figure}[!t]
    \centering
    \resizebox{0.99\linewidth}{!}
    {
    \includegraphics[scale=0.21,valign=t]{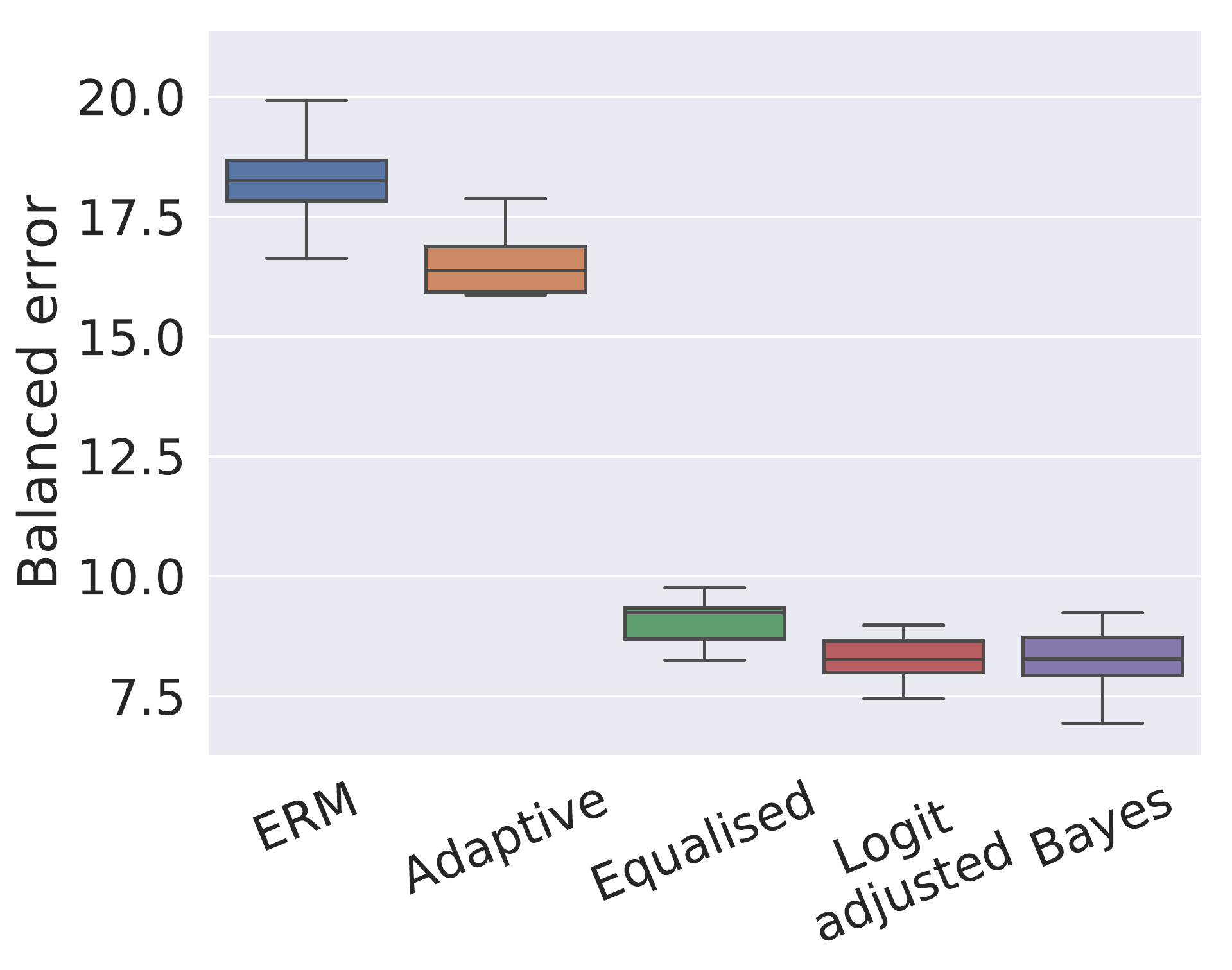}
    \quad%
    \includegraphics[scale=0.21,valign=t]{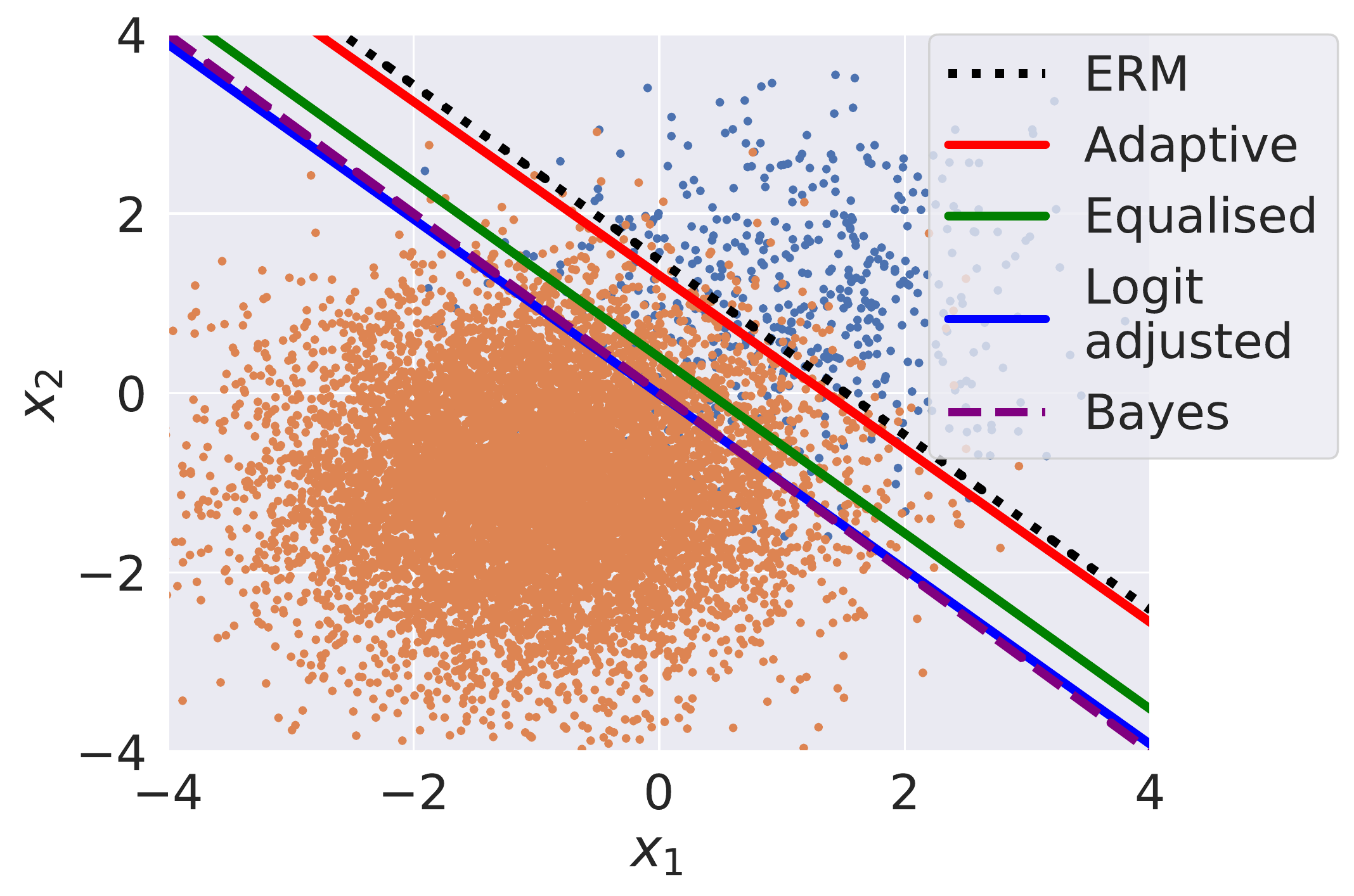}
    \quad%
    \includegraphics[scale=0.21,valign=t]{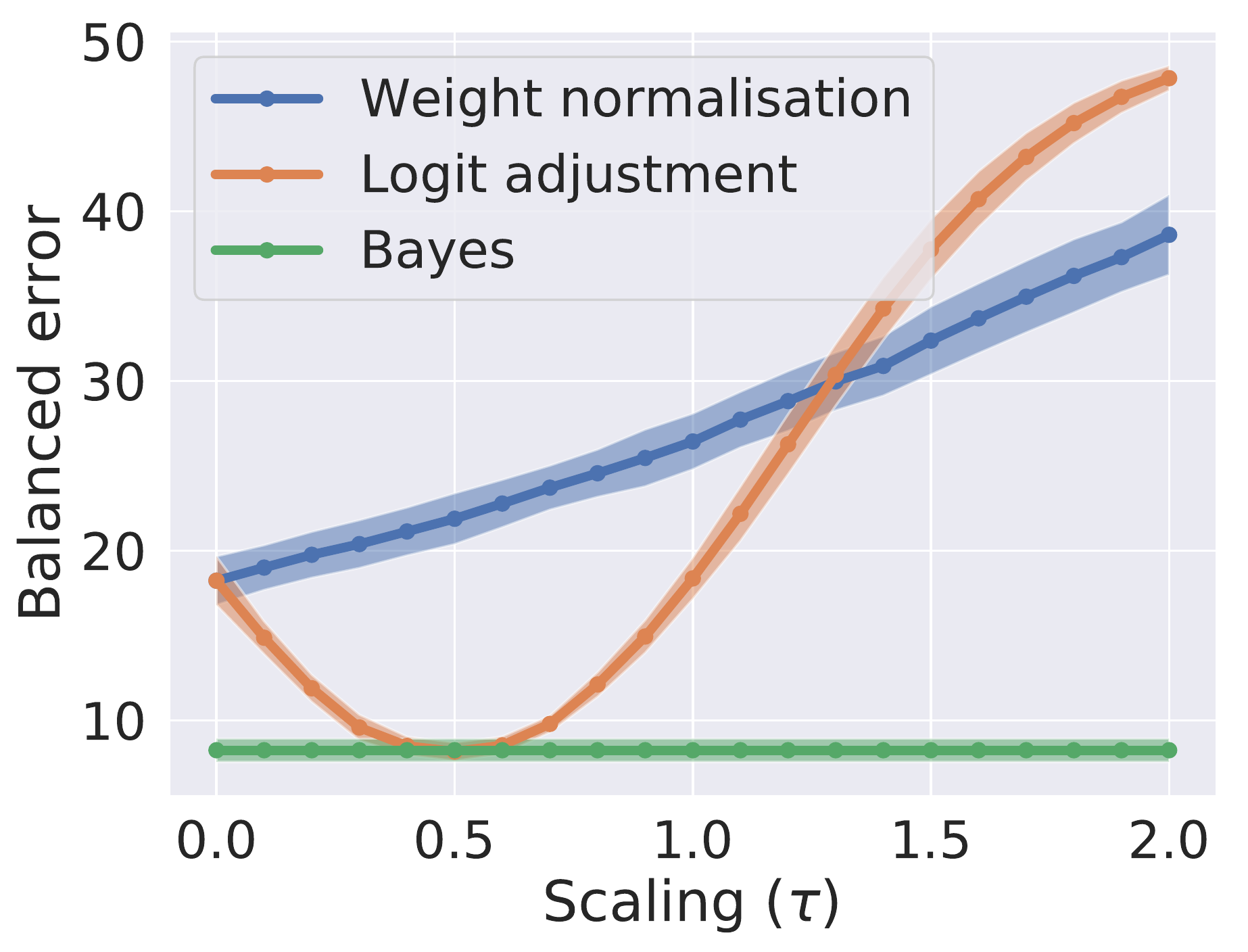}
    }%

    \caption{
    Results on synthetic binary classification problem.
    Our logit adjusted loss tracks the Bayes-optimal solution
    and separator (left \& middle panel).
    Post-hoc logit adjustment matches the Bayes performance
    with suitable scaling (right panel);
    however, \emph{any} weight normalisation fails.
    }
    \label{fig:lt_synthetic}
    \vspace{-0.5\baselineskip}
\end{figure}

Figure~\ref{fig:lt_synthetic} confirms that 
the logit adjusted margin loss attains a balanced error close to that of the Bayes-optimal,
which is visually reflected by its learned separator closely matching that in~\eqref{eqn:bayes-gaussian}.
This is in line with our claim of the logit adjusted margin loss being consistent for the balanced error, unlike other approaches.
Figure~\ref{fig:lt_synthetic} also compares post-hoc weight normalisation and logit adjustment
for varying scaling parameter $\tau$ (c.f.~\eqref{eqn:weight-normalisation},~\eqref{eqn:logit-adjustment}).
Logit adjustment is seen to approach the performance of the Bayes predictor;
\emph{any} weight normalisation is however seen to hamper performance.
This verifies the consistency of logit adjustment,
and inconsistency of weight normalisation (\S\ref{sec:weight-norm-critique}).

%
\subsection{Results on real-world datasets}
\label{sec:experiments-real}

We present results on the CIFAR-10, CIFAR-100, ImageNet and iNaturalist 2018 datasets.
Following prior work,
we create ``long-tailed versions'' of the CIFAR datasets by suitably downsampling examples per label 
following the
\expp profile of~\citet{Cui:2019,Cao:2019} with imbalance ratio 
$\rho = {\max_{y} \Pr( y )} / {\min_{y} \Pr( y )} = 100$. 
Similarly, we use the long-tailed version of ImageNet produced by~\citet{Liu:2019}.
We employ a ResNet-32 for CIFAR,
and a ResNet-50 for ImageNet and iNaturalist.
All models are trained using SGD with momentum;
see Appendix~\ref{app:architectures} for more details.
See also Appendix~\ref{app:additional_results} for results on CIFAR under the {\sc Step} profile considered in the literature.

\begin{table}[!t]
    \centering
    \renewcommand{\arraystretch}{1.25}

    \scalebox{0.8}{
    \begin{tabular}{@{}lllll@{}}
        \toprule
        \textbf{Method} & \textbf{CIFAR-10-LT} & \textbf{CIFAR-100-LT} & \textbf{ImageNet-LT} & \textbf{iNaturalist} \\
        \toprule
        ERM                  & 
        27.16 & 
        61.64 &
        %
        53.11 &
        38.66 \\
        Weight normalisation ($\tau = 1$)~\citep{Kang:2020} & 
        24.02 & 
        %
        58.89 &
        %
        52.00 &
        48.05 \\ 
        Weight normalisation ($\tau = \tau^*$)~\citep{Kang:2020} & 
        21.50 & 
        58.76 &
        %
        49.37 & 
        %
        34.10$^\star$ \\ 
        Adaptive~\citep{Cao:2019}             & 
        26.65$^\dagger$ & 
        60.40$^\dagger$ & 
        %
        52.15 &
        35.42$^\dagger$ \\
        Equalised~\citep{Tan:2020}            & 
        26.02 & 
        %
        57.26 & 
        %
        54.02 &
        %
        38.37 \\
        \midrule
        Logit adjustment post-hoc ($\tau = 1$) & 
        22.60 & 
        %
        58.24 &
        %
        49.66 &
        33.98 \\
        Logit adjustment post-hoc ($\tau = \tau^*$) & 
        \best{19.08} & 
        57.90 & 
        49.56 &
        33.80 \\
        Logit adjustment loss ($\tau = 1$)    & 
        {22.33}      & 
        \best{56.11} & 
        \best{48.89} &
        \best{33.64} \\
        \bottomrule
    \end{tabular}
    }

    \caption{
    Test set balanced error (averaged over $5$ trials) on real-world datasets.
    \arxiv{}{We use a ResNet-32 for the CIFAR datasets, and ResNet-50 for the ImageNet and iNaturalist datasets.}
    Here, $^\dagger$, $^\star$ are numbers for ``LDAM + SGD'' from~\citet[Table 2, 3]{Cao:2019}
    and ``$\tau$-normalised'' from~\citet[Table 3, 7]{Kang:2020}.
    \arxiv{See Figure~\ref{fig:cifar100_post_hoc_comparison} for results when $\tau \neq 1$, and discussion about further extensions.}{%
    Here, $\tau = \tau^*$ refers to using the best possible value of tuning parameter $\tau$.
    See Figure~\ref{fig:cifar100_post_hoc_comparison} for plots as a function of $\tau$, and the ``Discussion'' subsection for further extensions.}
    }
    \label{tbl:results}
    \arxiv{\vspace{-\baselineskip}}{}
\end{table}

\textbf{Baselines}.
We consider:
\begin{enumerate*}[label=(\roman*)]
    \item empirical risk minimisation (ERM) on the long-tailed data,

    \item post-hoc weight normalisation~\citep{Kang:2020}
per \eqref{eqn:weight-normalisation} (using $\nu_y = \| w_y \|_2$ and $\tau = 1$) 
applied to ERM,

    \item the adaptive margin loss~\citep{Cao:2019}
    per~\eqref{eqn:cao},
    and

    \item 
    the equalised loss~\citep{Tan:2020}
    per~\eqref{eqn:equalised},
    with 
    $\delta_{y'} = F( \pi_{y'} )$
    for the threshold-based $F$ of~\citet{Tan:2020}.
\end{enumerate*}
\citet{Cao:2019} demonstrated superior performance of their adaptive margin loss against several other baselines,
such as the balanced loss of~\eqref{eqn:balanced-loss},
and that of~\citet{Cui:2019}.
Where possible, we report numbers for the baselines 
(which use the same setup as above)
from the respective papers.
See also our concluding discussion about extensions to such methods that improve performance.

We compare the above methods against
our proposed post-hoc logit adjustment~\eqref{eqn:logit-adjustment},
and logit adjusted loss~\eqref{eqn:logit-adjusted-loss}.
For 
post-hoc logit adjustment, we fix the scalar $\tau = 1$;
we analyse the effect of tuning this in Figure~\ref{fig:cifar100_post_hoc_comparison}.
We do not perform \emph{any} further tuning of our logit adjustment techniques.

\textbf{Results and analysis}.
Table~\ref{tbl:results} summarises our results,
which demonstrate
our proposed logit adjustment techniques consistently outperform existing methods.
\newedit{
Indeed,
while weight normalisation offers gains over ERM, these are improved significantly by post-hoc logit adjustment (e.g., 8\% relative reduction on CIFAR-10).
Similarly loss correction techniques are generally outperformed by our logit adjusted softmax cross-entropy (e.g., 6\% relative reduction on iNaturalist).

%
\begin{figure}[!t]
    \centering

    \subcaptionbox{CIFAR-10.}{\includegraphics[scale=0.21]{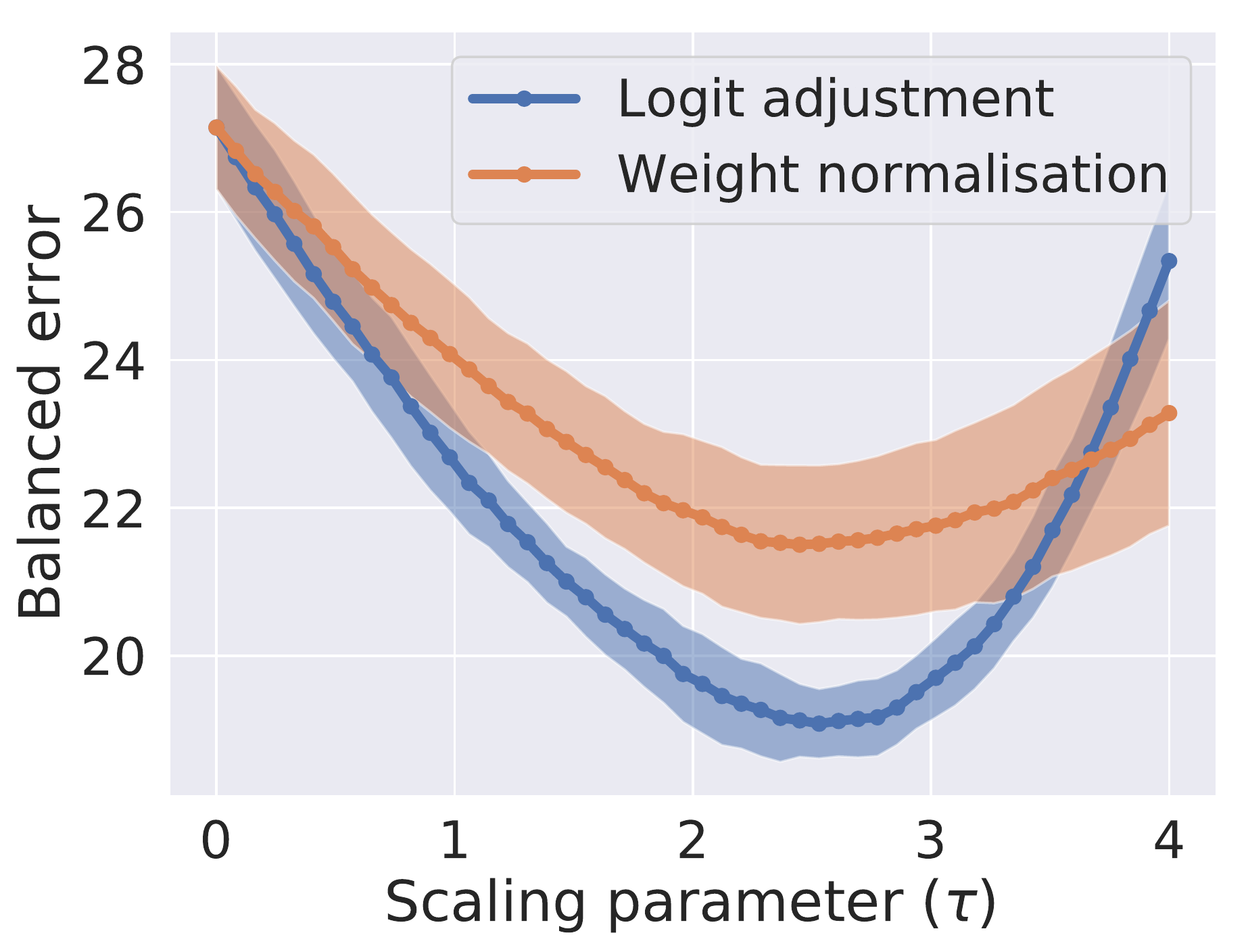}}%
    \quad%
    \subcaptionbox{CIFAR-100.}{\includegraphics[scale=0.21]{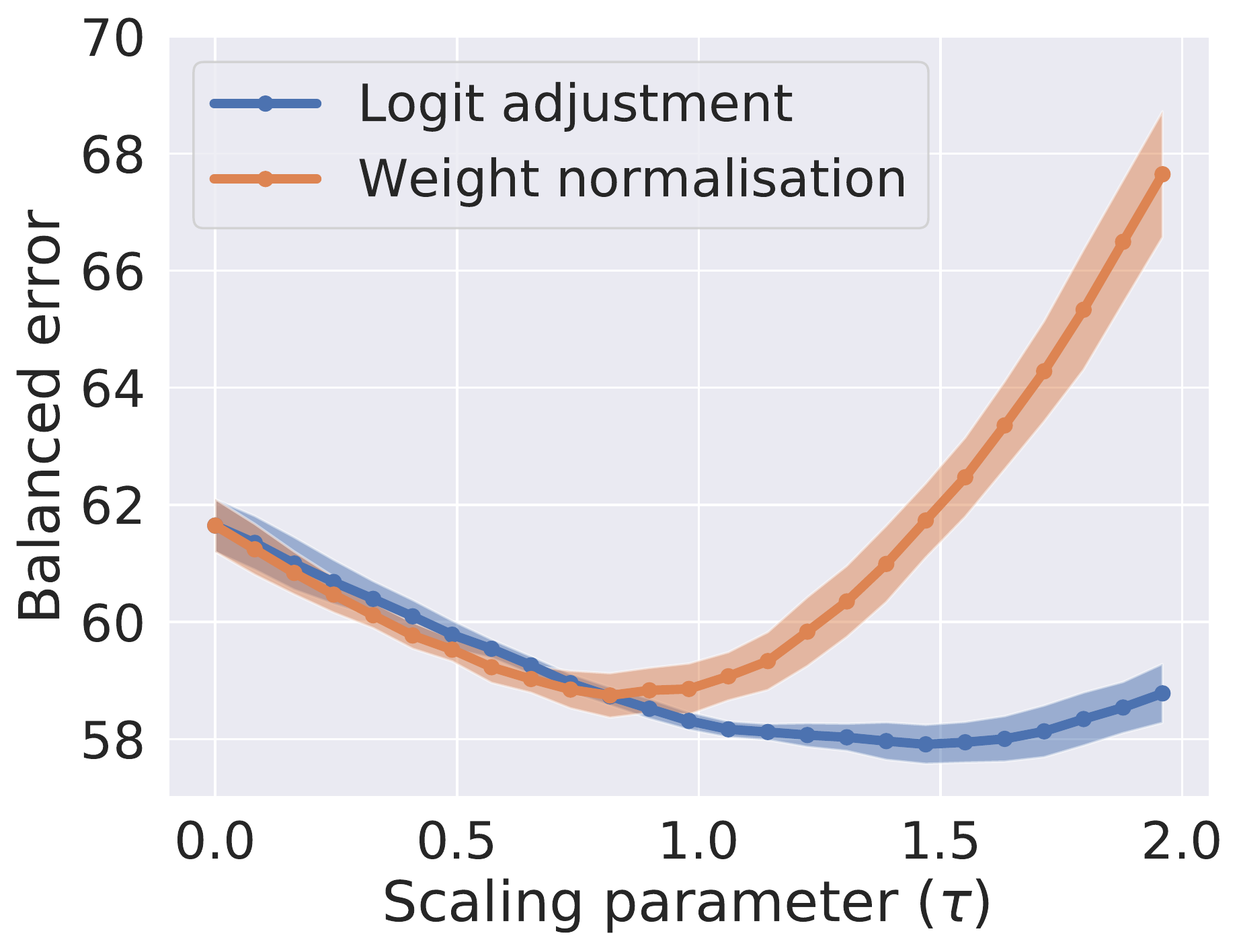}}%
    \quad%
    \subcaptionbox{iNaturalist.}{\includegraphics[scale=0.21]{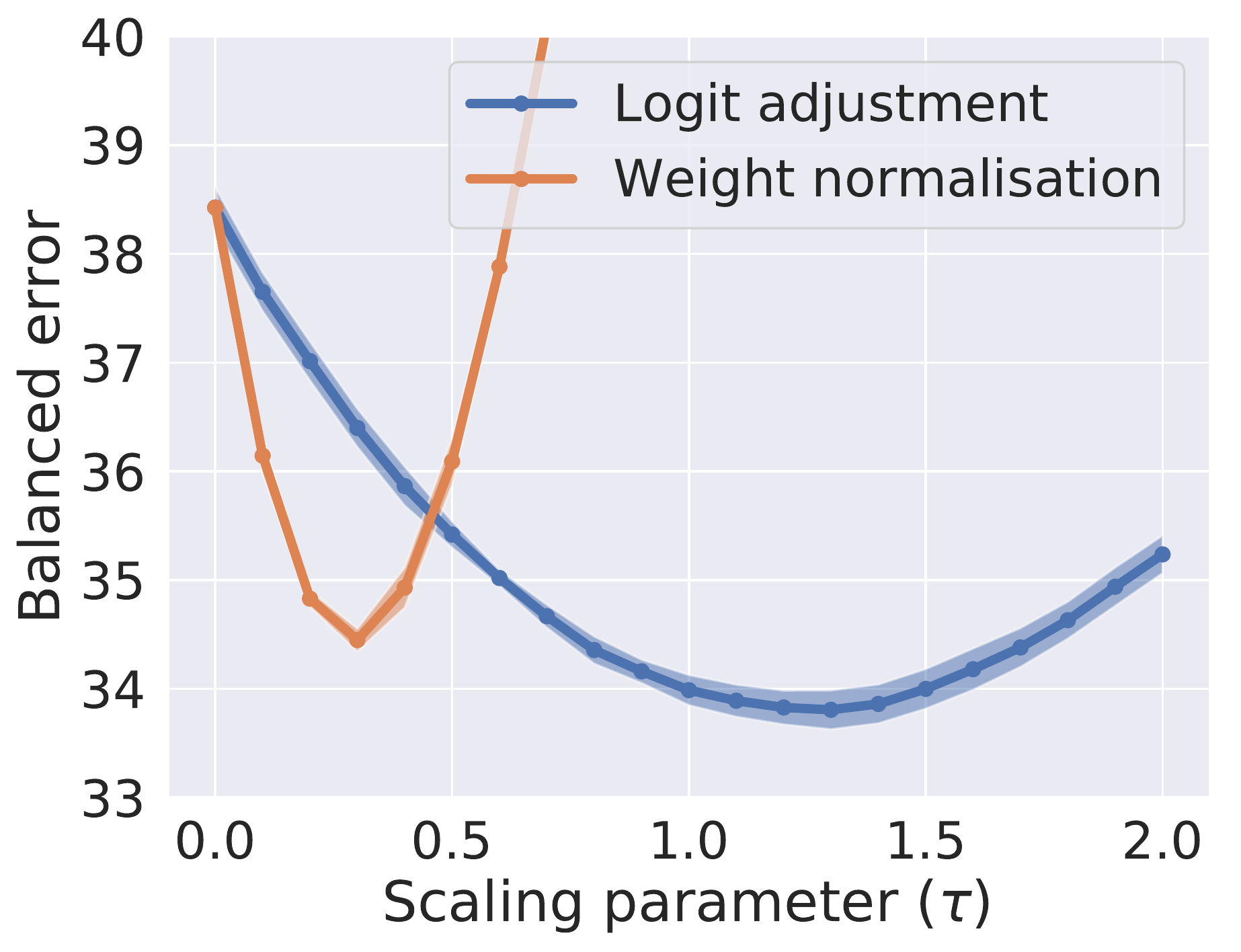}}

    \caption{
    Comparison of balanced error
    for post-hoc correction techniques    
    when varying scaling parameter $\tau$ (c.f.~\eqref{eqn:weight-normalisation},~\eqref{eqn:logit-adjustment}).
    Post-hoc logit adjustment consistently outperforms weight normalisation.
    }
    \label{fig:cifar100_post_hoc_comparison}
    \arxiv{\vspace{-0.5\baselineskip}}{}
\end{figure}

\begin{figure}
    \centering
    \subcaptionbox{CIFAR-10.}{\includegraphics[scale=0.1875,valign=t]{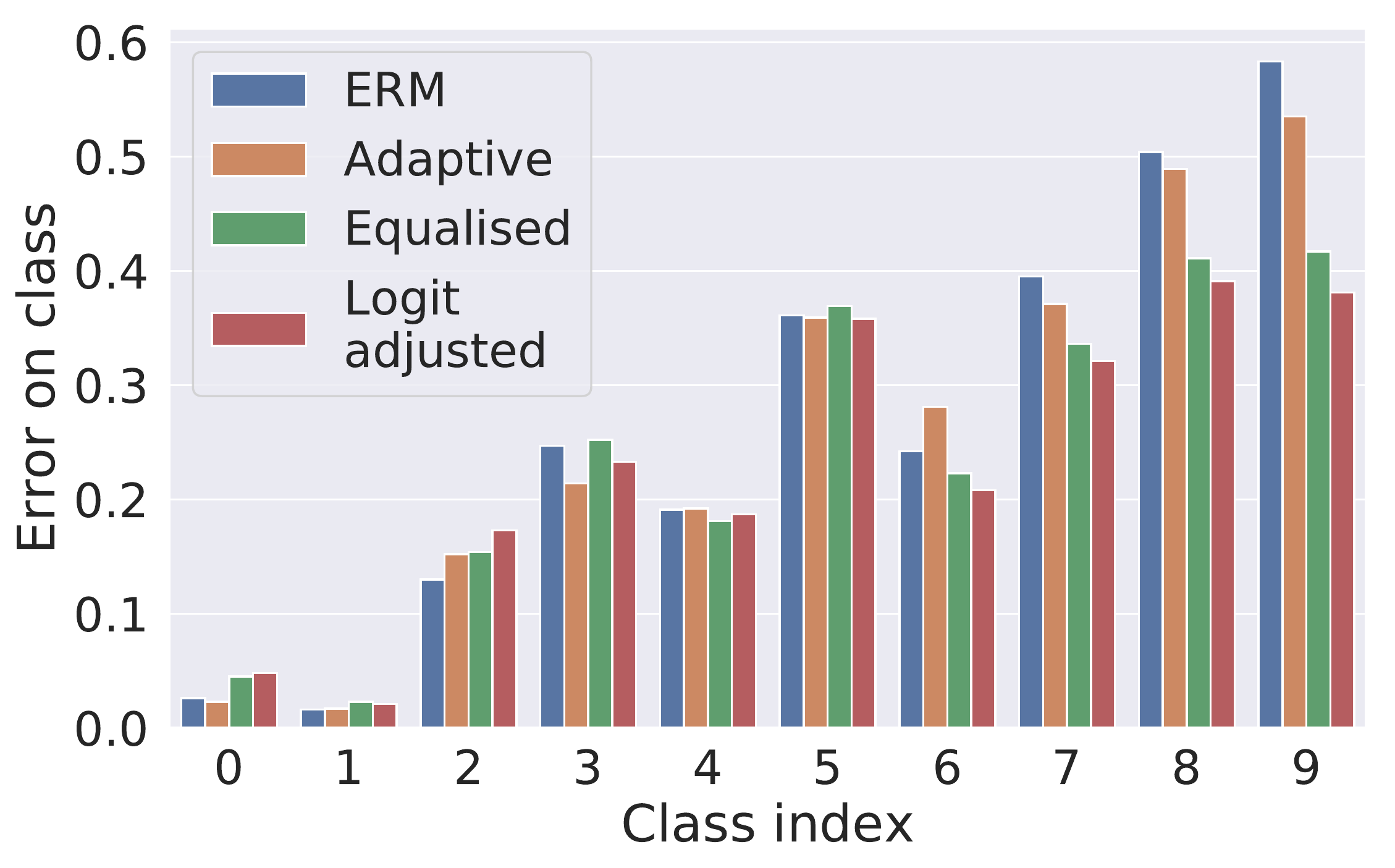}}%
    \quad%
    \subcaptionbox{CIFAR-100.}{\includegraphics[scale=0.1875,valign=t]{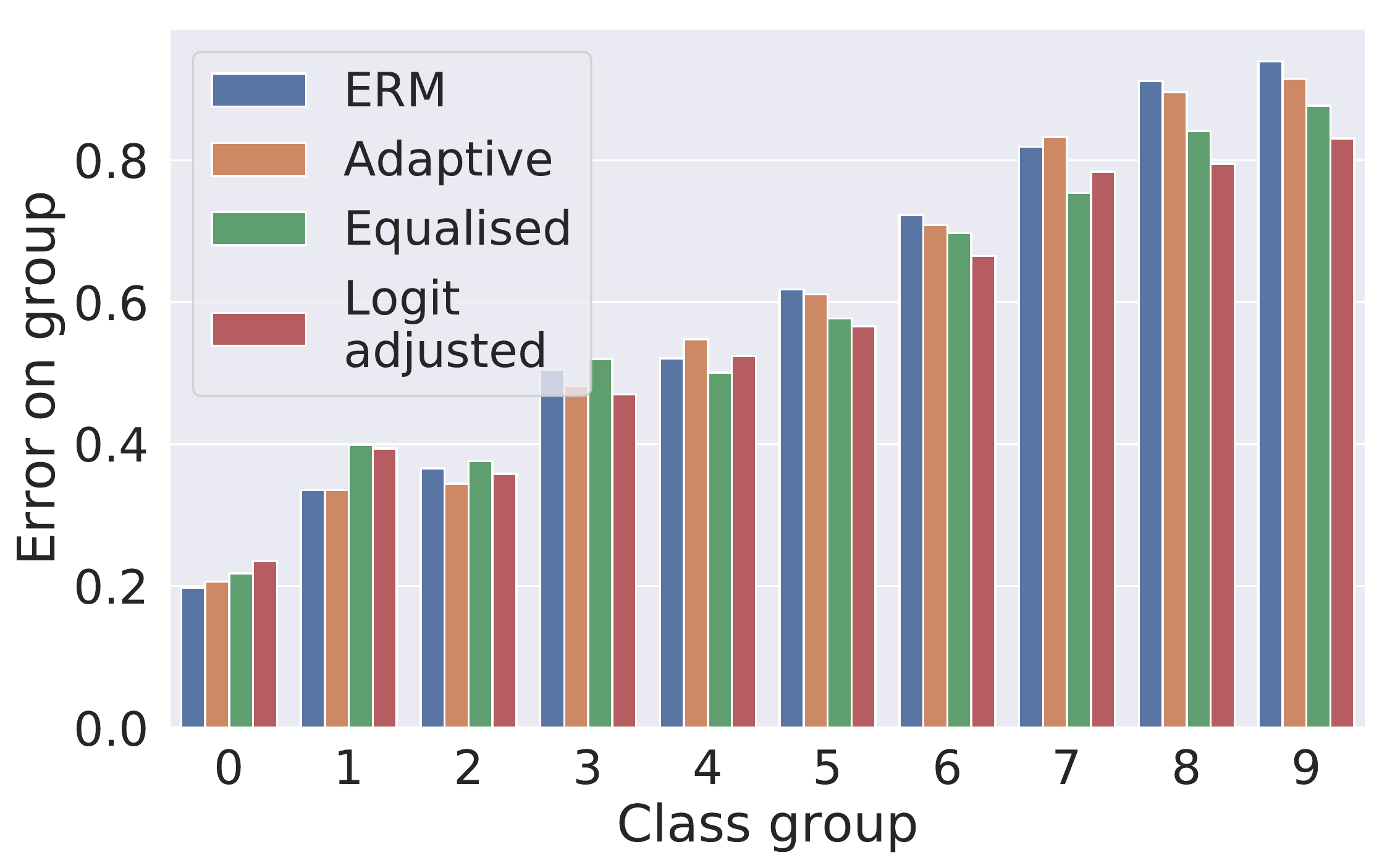}}
    \quad%
    \subcaptionbox{iNaturalist.}{\includegraphics[scale=0.1875,valign=t]{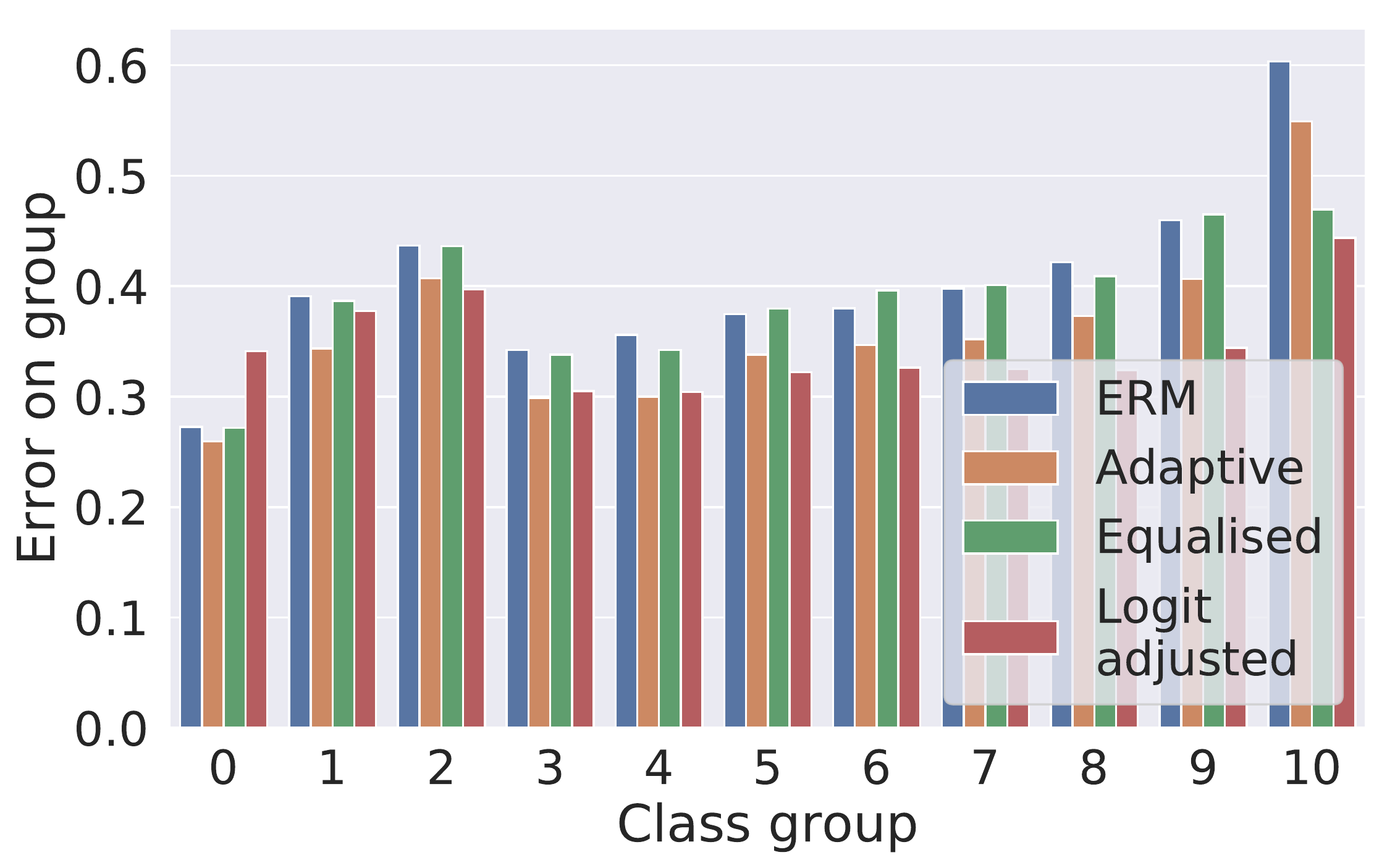}}    

    \caption{Per-class error rates of loss modification techniques. 
    For (b) and (c), we aggregate the classes into 10 groups.
    ERM displays a strong bias towards dominant classes (lower indices).
    Our proposed logit adjusted softmax loss achieves significant gains on rare classes (higher indices). 
    }
    \label{fig:per_class_accuracies}
    \arxiv{\vspace{-1.5\baselineskip}}{}
\end{figure}

%
Figure~\ref{fig:cifar100_post_hoc_comparison}  studies the effect of tuning the scaling parameter $\tau > 0$
afforded by post-hoc weight normalisation (using $\nu_y = \| w_y \|_2$) and post-hoc logit adjustment.
Even without \emph{any} scaling, post-hoc logit adjustment generally offers superior performance to the best result from weight normalisation {(cf.~Table~\ref{tbl:results})};
with scaling, this is further improved.
See Appendix~\ref{sec:additional_experiments} for a plot on ImageNet-LT.

Figure~\ref{fig:per_class_accuracies} breaks down the per-class accuracies on CIFAR-10, CIFAR-100, and iNaturalist.
On the latter two datasets, for ease of visualisation,
we aggregate the classes into ten groups based on their frequency-sorted order (so that, e.g., group $0$ comprises the top $\frac{L}{10}$ most frequent classes).
As expected,
dominant classes generally see a lower error rate with all methods.
However,
the logit adjusted loss is seen to systematically improve performance over ERM, particularly on rare classes.

While our logit adjustment techniques perform similarly,
there is a slight advantage to the loss function version.
Nonetheless, the strong performance of post-hoc logit adjustment corroborates the ability to decouple representation and classifier learning in long-tail settings~\citep{Zhang:2019}.

\arxiv{\textbf{Discussion and future work}.}{\textbf{Discussion and extensions}}
Table~\ref{tbl:results} shows the advantage of logit adjustment over recent post-hoc and loss modification proposals,
under standard setups from the literature.
We believe further improvements are possible by fusing complementary ideas, and remark on four such options.

First,
one may use a more complex base architecture;
our choices are standard in the literature,
but, e.g.,~\citet{Kang:2020} found gains on ImageNet-LT by employing a ResNet-152,
with further gains from training it for $200$ as opposed to the customary $90$ epochs.
\arxiv{Given the results in Figure~\ref{fig:cifar100_post_hoc_comparison}, we believe logit adjustment can similarly see gains.}{%
Table~\ref{tbl:results_architecture} confirms that logit adjustment similarly benefits from this choice.
For example, on iNaturalist, we obtain an improved balanced error of {${31.15}\%$} for the logit adjusted loss.
When training for more ($200$) epochs per the suggestion of~\citet{Kang:2020}, this further improves to {${30.12}\%$}.

{
Second, one may combine together the $\Delta$'s for various special cases of the pairwise margin loss. 
Indeed, we find that combining our relative margin with the adaptive margin of~\citet{Cao:2019} 
---
i.e., using the pairwise margin loss with $\Delta_{yy'} = \log \frac{\pi_{y'}}{\pi_{y}} + \frac{1}{\pi_{y}^{1/4}}$
---
results in a top-$1$ accuracy of {${31.56}\%$} on iNaturalist.
When using a ResNet-152, this further improves to {${29.22}\%$} when trained for 90 epochs,
and {$\mathbf{28.02}\%$} when trained for 200 epochs.
While such a combination is nominally heuristic, we believe there is scope to formally study such schemes, e.g., in terms of induced generalisation performance.
}

\begin{table}[!t]
    \centering
    \renewcommand{\arraystretch}{1.25}

    \resizebox{\linewidth}{!}{
    \begin{tabular}{@{}llllll@{}}
        \toprule
        & \multicolumn{2}{c}{\bf ImageNet-LT} & \multicolumn{2}{c}{\bf iNaturalist} \\
        \textbf{Method} & \textbf{ResNet-50} & \textbf{ResNet-152} & \textbf{ResNet-50} & \textbf{ResNet-152} & \textbf{ResNet-152} \\
        \textbf{} & \textbf{} & \textbf{} & \textbf{90 epochs} & \textbf{90 epochs} & \textbf{200 epochs} \\
        \toprule
        ERM                  & 
        53.11 &  
        53.30 &
        38.66 & 
        35.88 & 
        34.38 \\
        Weight normalisation ($\tau = 1$)~\citep{Kang:2020} & 
        52.00 & 
        51.49 &
        48.05 & 
        45.17 & 
        45.33 \\ 
        Weight normalisation ($\tau = \tau^*$)~\citep{Kang:2020} & 
        49.37 &
        48.97 &
        34.10 & 
        31.85 & 
        30.34 \\ 
        Adaptive~\citep{Cao:2019}             & 
        52.15 & 
        53.34 &
        35.42 & 
        31.18 & 
        29.46 \\
        Equalised~\citep{Tan:2020}            & 
        54.02 & 
        51.38 &
        38.37 & 
        35.86 & 
        34.53 \\
        \midrule
        Logit adjustment post-hoc ($\tau = 1$) & 
        49.66 & 
        49.25 &
        33.98 & 
        31.46 & 
        30.15 \\
        Logit adjustment post-hoc ($\tau = \tau^*$) & 
        49.56 & 
        49.15 &
        33.80 & 
        31.08 & 
        29.74 \\
        Logit adjustment loss ($\tau = 1$)    & 
        \best{48.89} & 
        \best{47.86} &
        {33.64} & 
        {31.15} & 
        30.12 \\
        Logit adjustment plus adaptive loss ($\tau = 1$)    & 
        {51.25} & 
        {50.46} & 
        \best{31.56} & 
        \best{29.22} & 
        \best{28.02} \\
        \bottomrule
    \end{tabular}
    }

    \caption{
    Test set balanced error (averaged over $5$ trials) on real-world datasets with more complex base architectures.
    Employing a ResNet-152 is seen to systematically improve all methods' performance, with logit adjustment remaining superior to existing approaches.
    The final row reports the results of combining logit adjustment with the adaptive margin loss of~\citet{Cao:2019}, which yields further gains on iNaturalist.
    }
    \label{tbl:results_architecture}
    \arxiv{\vspace{-\baselineskip}}{}
\end{table}
}

{Third,}
\citet{Cao:2019} observed that their loss benefits from a deferred reweighting scheme (DRW), wherein the model 
begins training
as normal,
and then
applies class-weighting
after a fixed number of epochs.
On CIFAR-10-LT and CIFAR-100-LT,
this 
achieves $22.97\%$ and $57.96\%$
error respectively;
both are outperformed by
our vanilla logit adjusted loss.
On iNaturalist\arxiv{}{ with a ResNet-50}, 
this
achieves an error of $32.0\%$, outperforming our $33.6\%$.
\arxiv{However, given the strong improvement of our loss over that in~\citet{Cao:2019} when both methods use SGD,
we expect that employing DRW 
(which applies to any loss)
can be similarly beneficial for our method.}
{(Note that our simple combination of the relative and adaptive margins outperforms these reported numbers of DRW.)
However, given the strong improvement of our loss over that in~\citet{Cao:2019} when both methods use SGD,
we expect that employing DRW 
(which applies to any loss)
may be similarly beneficial for our method.}

\arxiv{Second,}{Fourth,} per~\S\ref{sec:background}, one may perform data augmentation;
e.g., see~\citet[Section 6]{Tan:2020}.
While further exploring such variants are of empirical interest,
we hope to have illustrated the conceptual and empirical value of logit adjustment,
and leave this for future work.
}


\clearpage

\arxiv{%
\section*{Broader impact}
Our work presents new analysis and algorithms for the foundational learning problem of supervised classification where certain classes are rare.
\newedit{This problem is important, e.g., to ensure fairness in machine learning when certain data subpopulations are under-represented.}
The benefits of this work are that it allows for better performance of classification models on classes which do not have many associated samples.
When our methods do not work, it would result in poorer performance on such classes;
\newedit{however, this is still expected to be superior to na\"{i}ve training on the raw data}.
Our approaches do not explicitly model the data generation process, but do attempt to mitigate scenarios where certain classes are under-represented owing to bias in the training sample generation.

}{}

\bibliographystyle{plainnat}
\bibliography{references}

\clearpage
\begin{center}
    {\bf\Large Supplementary material for ``Long tail learning via logit adjustment''}
\end{center}

\appendix

\section{Proofs of results in body}
\begin{proof}[Proof of Theorem~\ref{thm:multiclass-unified-consistent}]
Denote $\eta_y( x ) = \Pr( y \mid x )$.
Suppose we employ a margin 
$\Delta_{y y'} = \log \frac{\delta_{y'}}{\delta_{y}}$.
Then, the loss is 
$$ {\ell}( y, f( x ) ) 
= -\log \frac{\delta_{y} \cdot e^{f_{y}( x )}}{\sum_{y' \in [L]} {\delta_{y'}} \cdot e^{f_{y'}( x )}}
= -\log \frac{e^{f_{y}( x ) + \log \delta_{y}}}{\sum_{y' \in [L]} e^{f_{y'}( x ) + \log \delta_{y'}}}. $$
Consequently, under constant weights $\alpha_y = 1$,
the Bayes-optimal score will satisfy
$f^*_y( x ) + \log \delta_{y} = \log \eta_y( x )$,
or
$f^*_y( x ) = \log \frac{\eta_y( x )}{\delta_y}$.

Now suppose we use generic weights $\alpha \in \Real_+^L$.
The risk under this loss is
\begin{align*}
    \E{\X, \Y}{ \ell_{\alpha}( \Y, f( \X ) ) } &= \sum_{y \in [L]} \pi_y \cdot \E{\X \mid \Y = y}{ \ell_{\alpha}( y, f( x ) ) } \\
    &= \sum_{y \in [L]} \pi_y \cdot \E{\X \mid \Y = y}{ \ell_{\alpha}( y, f( x ) ) } \\
    &= \sum_{y \in [L]} \pi_y \cdot \alpha_y \cdot \E{\X \mid \Y = y}{ \ell( y, f( x ) ) } \\
    &\propto \sum_{y \in [L]} \bar{\pi}_y \cdot \E{\X \mid \Y = y}{ \ell( y, f( x ) ) },
\end{align*}
where
$\bar{\pi}_y \propto {\pi_y \cdot \alpha_y}$.
Consequently, learning with the weighted loss is equivalent to learning with the original loss,
on a distribution with modified base-rates $\bar{\pi}$.
Under such a distribution, we have class-conditional distribution
$$ \bar{\eta}_y( x ) = \bar{\Pr}( y \mid x ) = \frac{ \Pr( x \mid y ) \cdot \bar{\pi}_y }{ \bar{\Pr}( x ) } = \eta_y( x ) \cdot \frac{\bar{\pi}_y}{{\pi}_y} \cdot \frac{\Pr( x )}{\bar{\Pr}( x )} \propto \eta_y( x ) \cdot \alpha_y. $$
Consequently, suppose
$\alpha_y = \frac{\delta_y}{\pi_y}$.
Then,
$f^*_y( x ) = \log \frac{\bar{\eta}_y( x )}{\delta_y} = \log \frac{\eta_y( x )}{\pi_y} + C( x )$,
where $C( x )$ does not depend on $y$.
Consequently,
$\argmaxUnique{y \in [L]}{f^*_y( x )} = \argmaxUnique{y \in [L]}{\frac{\eta_y( x )}{\pi_y}}$,
which is the Bayes-optimal prediction for the balanced error.

In sum, a consistent family can be obtained by choosing any set of constants $\delta_y > 0$ and setting
    \begin{align*}
        \alpha_y &= \frac{\delta_y}{\pi_y} \\
        \Delta_{y y'} &= \log \frac{\delta_{y'}}{\delta_{y}}.
    \end{align*}
\end{proof}


%
\section{On the consistency of binary margin-based losses}
It is instructive to study the pairwise margin loss~\eqref{eqn:unified-margin-loss} in the binary case.
Endowing the loss with a temperature parameter $\gamma > 0$, we get\footnote{Compared to the multiclass case, we assume here a scalar score $f \in \Real$. This is equivalent to constraining that $\sum_{y \in [L]} f_{y} = 0$ for the multiclass case.}
\begin{equation}
    \label{eqn:binary-unified}
    \begin{aligned}
        \ell( +1, f ) &= \frac{\omega_{+1}}{\gamma} \cdot \log( 1 + e^{\gamma \cdot \delta_{+1}} \cdot e^{-\gamma \cdot f} ) \\
        \ell( -1, f ) &= \frac{\omega_{-1}}{\gamma} \cdot \log( 1 + e^{\gamma \cdot \delta_{-1}} \cdot e^{\gamma \cdot f} )
    \end{aligned}
\end{equation}
for constants $\omega_{\pm 1}, \gamma > 0$
and $\delta_{\pm 1} \in \Real$.
Here, we have used $\delta_{+1} = \Delta_{+1, -1}$ and $\delta_{-1} = \Delta_{-1, +1}$ for simplicity.
The choice $\omega_{\pm 1} = 1, \delta_{\pm 1} = 0$ recovers the temperature scaled binary logistic loss.
Evidently, as $\gamma \to +\infty$, these converge to weighted hinge losses with variable margins,
i.e.,
\begin{align*}
    \ell( +1, f ) &= {\omega_{+1}} \cdot [ \delta_{+1} - f ]_+ \\
    \ell( -1, f ) &= {\omega_{-1}} \cdot [ \delta_{-1} + f ]_+.
\end{align*}

We study two properties of this family losses.
First, under what conditions are the losses Fisher consistent for the balanced error?
We shall show that in fact there is a simple condition characterising this.
Second, do the losses preserve properness of the original binary logistic loss?
We shall show that this is always the case, but that the losses involve fundamentally different approximations.

\subsection{Consistency of the binary pairwise margin loss}
\label{sec:binary-consistency}

Given a loss $\ell$, its \emph{Bayes optimal} solution is $f^* \in \operatorname{argmin}_{f \colon \XCal \to \Real} \E{}{ \ell( \Y, f( \X ) ) }$.
For consistency with respect to the balanced error in the binary case, we require this optimal solution $f^*$ to satisfy
$f^*( x ) > 0 \iff {\eta( x ) > \pi}$,
where $\eta( x ) \defEq \Pr( y = 1 \mid x )$ and $\pi \defEq \Pr( y = 1 )$~\citep{Menon:2013}.
This is equivalent to a simple condition on the weights $\omega$ and margins $\delta$ of the pairwise margin loss.

\begin{lemma}
\label{lemm:binary-unified-consistent}
The losses in~\eqref{eqn:binary-unified} are consistent for the balanced error iff
$$ \frac{\omega_{+1}}{\omega_{-1}} \cdot \frac{\sigma( \gamma \cdot \delta_{+1} )}{\sigma( \gamma \cdot \delta_{-1} )} = \frac{1 - \pi}{\pi}, $$
where $\sigma( z ) = (1 + \exp( z ) )^{-1}$.
\end{lemma}

\begin{proof}[Proof of Lemma~\ref{lemm:binary-unified-consistent}]
Denote $\eta( x ) \defEq \Pr( y = +1 \mid x )$,
and $\pi \defEq \Pr( y = +1 )$.
From Lemma~\ref{lemm:binary-unified-proper} below, the pairwise margin loss is proper composite with invertible link function $\Psi \colon [0, 1] \to \Real \cup \{ \pm \infty \}$.
Consequently,
since by definition the Bayes-optimal score for a proper composite loss is $f^*( x ) = \Psi( \eta( x ) )$~\citep{Reid:2010},
to have consistency for the balanced error, 
from~\eqref{eqn:proper-link},~\eqref{eqn:binary-loss-derivative},
we require
\begin{align*}
    \Psi^{-1}( 0 ) = \pi &\iff \frac{1}{1 - \frac{\ell'( +1, 0 )}{\ell'( -1, 0 )}} = \pi \\
    &\iff {1 - \frac{\ell'( +1, 0 )}{\ell'( -1, 0 )}} = \frac{1}{\pi} \\
    &\iff {-\frac{\ell'( +1, 0 )}{\ell'( -1, 0 )}} = \frac{1 - \pi}{\pi} \\
    &\iff \frac{\omega_{+1}}{\omega_{-1}} \cdot \frac{\sigma( \gamma \cdot \delta_{+1} )}{\sigma( \gamma \cdot \delta_{-1} )} = \frac{1 - \pi}{\pi}.
\end{align*}
\end{proof}

From the above,
some admissible parameter choices include:
\begin{itemize}
    \item $\omega_{+1} = \frac{1}{\pi}$, $\omega_{-1} = \frac{1}{1 - \pi}$, 
    $\delta_{\pm 1} = 1$; 
    i.e., the standard weighted loss with a constant margin
    
    \item $\omega_{\pm 1} = 1$, 
    $\delta_{+1} = \frac{1}{\gamma} \cdot \log \frac{1 - \pi}{\pi}$, 
    $\delta_{-1} = \frac{1}{\gamma} \cdot \log \frac{\pi}{1 - \pi}$;
    i.e., the unweighted loss with a margin biased towards the rare class, as per our logit adjustment procedure
\end{itemize}
The second example above is unusual in that it requires scaling the margin with the temperature; consequently, the margin disappears as $\gamma \to +\infty$.
Other combinations are of course possible, but note that one cannot arbitrarily choose parameters and hope for consistency in general.
Indeed, some \emph{inadmissible} choices are na\"{i}ve applications of the margin modification or weighting,
e.g.,
\begin{itemize}
    \item $\omega_{+1} = \frac{1}{\pi}$, $\omega_{-1} = \frac{1}{1 - \pi}$, 
    $\delta_{+1} = \frac{1}{\gamma} \cdot \log \frac{1 - \pi}{\pi}$, 
    $\delta_{-1} = \frac{1}{\gamma} \cdot \log \frac{\pi}{1 - \pi}$;
    i.e., combining \emph{both} weighting and margin modification

    \item $\omega_{\pm 1} = 1$, $\delta_{+1} = \frac{1}{\gamma} \cdot (1 - \pi)$, $\delta_{-1} = \frac{1}{\gamma} \cdot {\pi}$;
    i.e., specific margin modification
\end{itemize}

Note further that the choices of~\citet{Cao:2019,Tan:2020} do not meet the requirements of Lemma~\ref{lemm:binary-unified-consistent}.

We make two final remarks.
First, the above only considers consistency of the result of loss minimisation.
For \emph{any} choice of weights and margins, we may apply suitable post-hoc correction to the predictions to account for any bias in the optimal scores.
Second, as $\gamma \to +\infty$, any \emph{constant} margins $\delta_{\pm 1} > 0$ will have no effect on the consistency condition, since $\sigma( \gamma \cdot \delta_{\pm 1} ) \to 1$.
The condition will be wholly determined by the weights $\omega_{\pm 1}$.
For example, we may choose
$\omega_{+1} = \frac{1}{\pi}$,
$\omega_{-1} = \frac{1}{1 - \pi}$,    
$\delta_{+1} = 1$,
and
$\delta_{-1} = \frac{\pi}{1-\pi}$;
the resulting loss will not be consistent for finite $\gamma$, but will become so in the limit $\gamma \to +\infty$.
For more discussion on this particular loss, see Appendix~\ref{app:cs-svm}.

\subsection{Properness of the pairwise margin loss}

In the above, we appealed to the pairwise margin loss being proper composite, in the sense of~\citet{Reid:2010}.
Intuitively, this specifies that the loss has Bayes-optimal score of the form $f^*( x ) = \Psi( \eta( x ) )$,
where $\Psi$ is some invertible function, and $\eta( x ) = \Pr( y = 1 \mid x )$.
We have the following general result about properness of \emph{any} member of the pairwise margin family.

\begin{lemma}
\label{lemm:binary-unified-proper}
The losses in~\eqref{eqn:binary-unified} are proper composite, with link function
\begin{align*}
    \Psi( p ) &= \frac{1}{\gamma} \cdot \log \left[ \left( \frac{a \cdot b}{q} - c \right) \pm \sqrt{\left( \frac{a \cdot b}{q} - c \right)^2 + 4 \cdot \frac{a}{q}} \right] - \log 2,
\end{align*}
where $a = \frac{\omega_{+1}}{\omega_{-1}} \cdot \frac{e^{\gamma \cdot \delta_{+1}}}{e^{\gamma \cdot \delta_{-1}}}$,
$b = e^{\gamma \cdot \delta_{-1}}$,
$c = e^{\gamma \cdot \delta_{+1}}$,
and $q = \frac{1 - p}{p}$.
\end{lemma}

\begin{proof}[Proof of Lemma~\ref{lemm:binary-unified-proper}]
The above family of losses is proper composite
iff the function
\begin{equation}
    \label{eqn:proper-link}
    \Psi^{-1}( f ) = \frac{1}{1 - \frac{\ell'( +1, f )}{\ell'( -1, f )}}
\end{equation}
is invertible~\citep[Corollary 12]{Reid:2010}.
We have
\begin{equation}
    \label{eqn:binary-loss-derivative}
    \begin{aligned}
        \ell'( +1, f ) &= -{\omega_{+1}} \cdot \frac{e^{\gamma \cdot \delta_{+1}} \cdot e^{-\gamma \cdot f}}{ 1 + e^{\gamma \cdot \delta_{+1}} \cdot e^{-\gamma \cdot f} } \\
        \ell'( -1, f ) &= +{\omega_{-1}} \cdot \frac{e^{\gamma \cdot \delta_{-1}} \cdot e^{\gamma \cdot f}}{ 1 + e^{\gamma \cdot \delta_{-1}} \cdot e^{\gamma \cdot f} }.
    \end{aligned}
\end{equation}
The invertibility of $\Psi^{-1}$ is immediate.
To compute the link function $\Psi$, note that
\begin{align*}
    p = \frac{1}{1 - \frac{\ell'( +1, f )}{\ell'( -1, f )}} &\iff \frac{1}{p} = 1 - \frac{\ell'( +1, f )}{\ell'( -1, f )} \\
    &\iff - \frac{\ell'( +1, f )}{\ell'( -1, f )} = \frac{1 - p}{p} \\
    &\iff \frac{\omega_{+1}}{\omega_{-1}} \cdot \frac{e^{\gamma \cdot \delta_{+1}} \cdot e^{-\gamma \cdot f}}{ 1 + e^{\gamma \cdot \delta_{+1}} \cdot e^{-\gamma \cdot f} } \cdot \frac{ 1 + e^{\gamma \cdot \delta_{-1}} \cdot e^{\gamma \cdot f} }{e^{\gamma \cdot \delta_{-1}} \cdot e^{\gamma \cdot f}} = \frac{1 - p}{p} \\
    &\iff \frac{\omega_{+1}}{\omega_{-1}} \cdot \frac{e^{\gamma \cdot \delta_{+1}}}{e^{\gamma \cdot \delta_{-1}}} \cdot \frac{1}{ e^{\gamma \cdot f} + e^{\gamma \cdot \delta_{+1}} } \cdot \frac{ 1 + e^{\gamma \cdot \delta_{-1}} \cdot e^{\gamma \cdot f} }{e^{\gamma \cdot f}} = \frac{1 - p}{p} \\
    &\iff a \cdot \frac{ 1 + b \cdot g  }{ g^2 + c \cdot g } = q,
\end{align*}
where $a = \frac{\omega_{+1}}{\omega_{-1}} \cdot \frac{e^{\gamma \cdot \delta_{+1}}}{e^{\gamma \cdot \delta_{-1}}}$,
$b = e^{\gamma \cdot \delta_{-1}}$,
$c = e^{\gamma \cdot \delta_{+1}}$,
$g = e^{\gamma \cdot f}$,
and $q = \frac{1 - p}{p}$.
Thus,
\begin{align*}
    a \cdot \frac{ 1 + b \cdot g  }{ g^2 + c \cdot g } = q &\iff \frac{ g^2 + c \cdot g }{ 1 + b \cdot g  } = \frac{a}{q} \\
    &\iff g^2 + \left( c - \frac{a \cdot b}{q} \right) \cdot g - \frac{a}{q} = 0 \\
    &\iff g = \frac{\left( \frac{a \cdot b}{q} - c \right) \pm \sqrt{\left( \frac{a \cdot b}{q} - c \right)^2 + 4 \cdot \frac{a}{q}}}{2}.
\end{align*}
\end{proof}

As a sanity check, suppose $a = b = c = \gamma = 1$.
This corresponds to the standard logistic loss.
Then,
$$ \Psi( p ) = \log \frac{\left( \frac{1}{q} - 1 \right) \pm \sqrt{\left( \frac{1}{q} - 1 \right)^2 + 4 \cdot \frac{1}{q}}}{2} = \log \frac{p}{1 - p}, $$
which is the standard logit function.

\begin{figure}[!htp]
    \centering
    \includegraphics[scale=0.275]{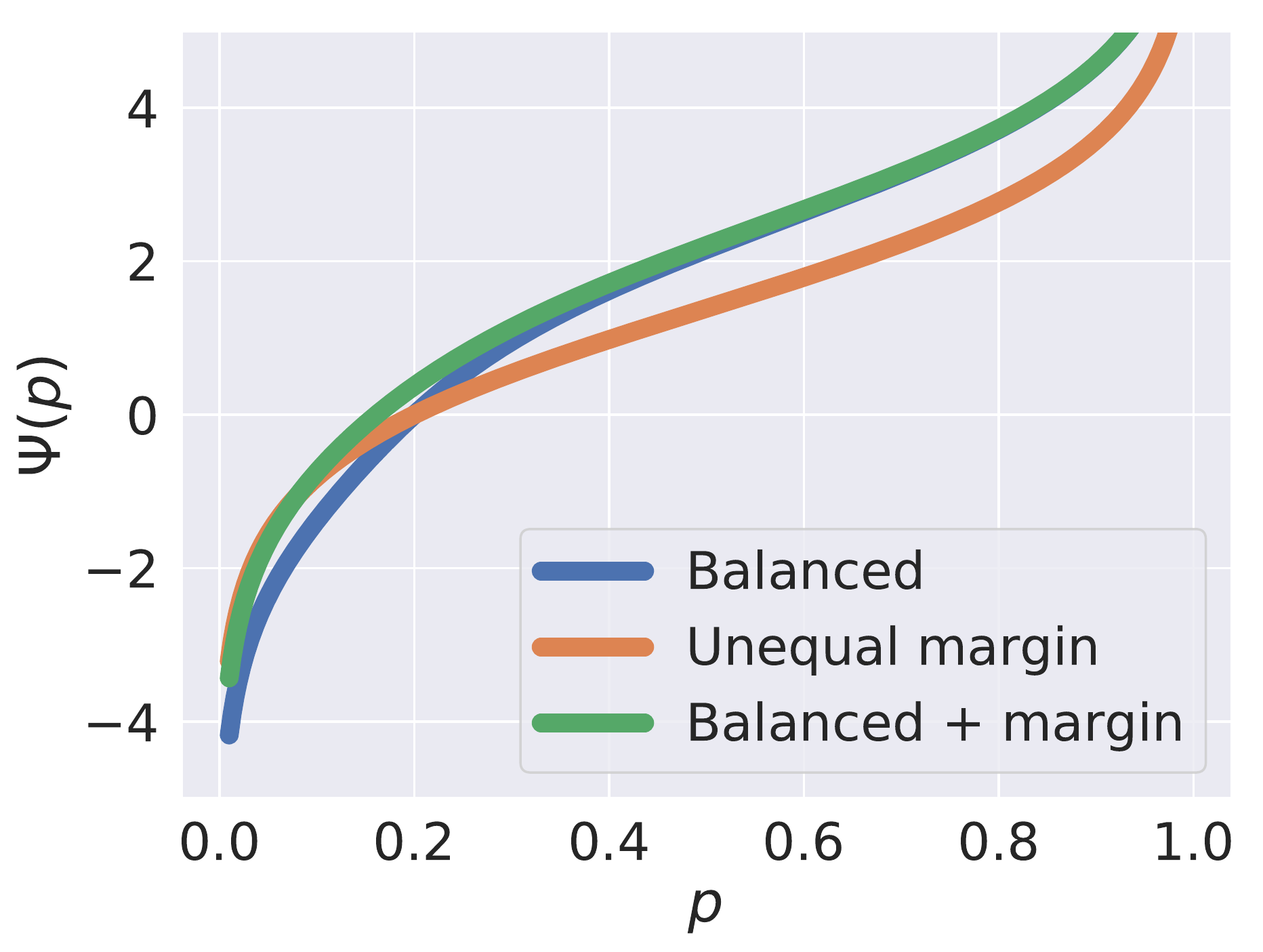}
    \includegraphics[scale=0.275]{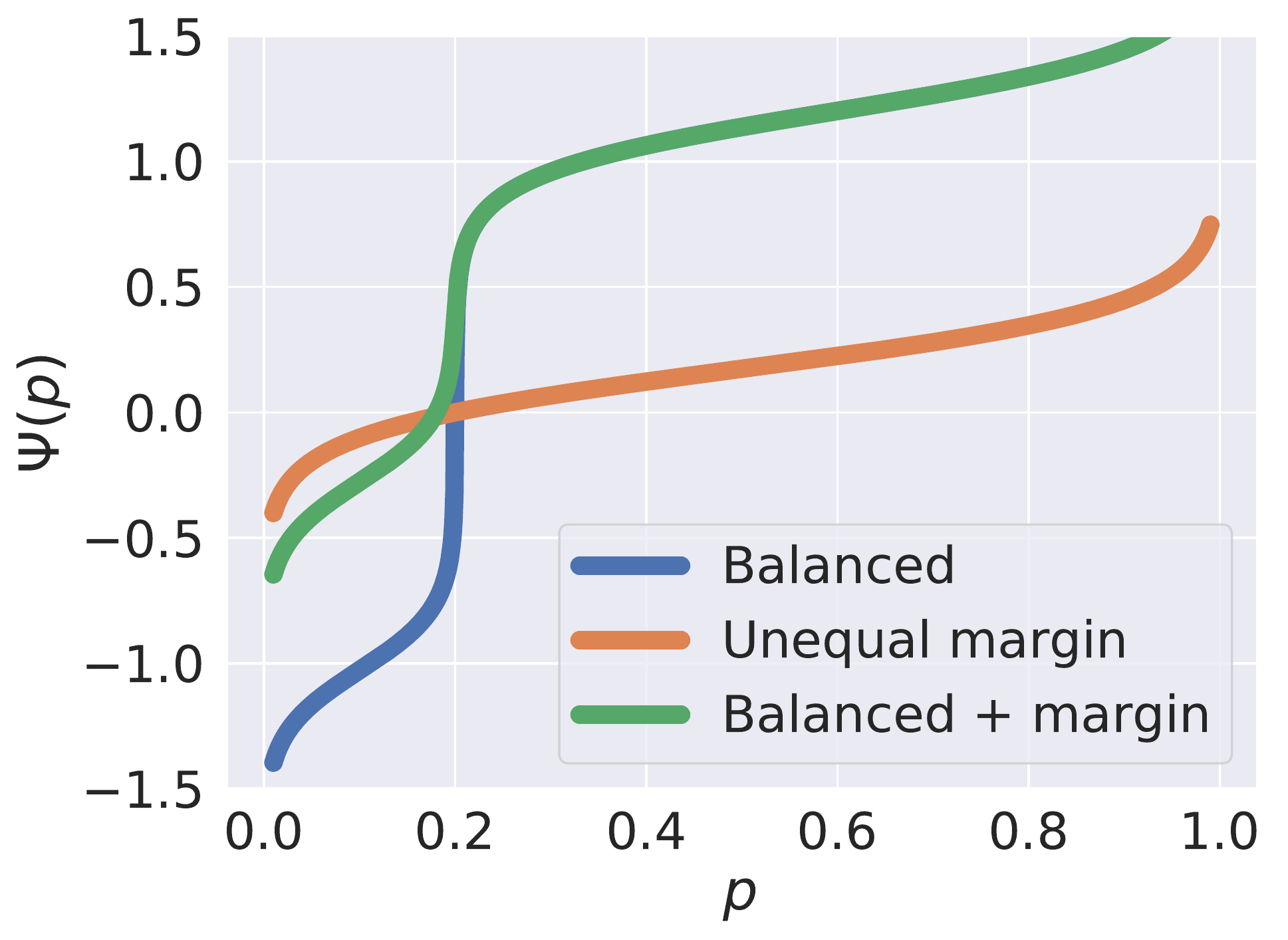}
    \caption{Comparison of link functions for various losses
    assuming $\pi = 0.2$,
    with $\gamma = 1$ (left) and $\gamma = 8$ (right).
    The balanced loss uses 
    $\omega_{y} = \frac{1}{\pi_{y}}$.
    The unequal margin loss uses 
    $\delta_{y} = \frac{1}{\gamma} \cdot \log \frac{1 - \pi}{\pi}$.
    The balanced + margin loss uses 
    $\delta_{-1} = \frac{\pi}{1-\pi}$, 
    $\delta_{+1} = 1$,
    $\omega_{+1} = \frac{1}{\pi}$.
    }
    \label{fig:link_comparison}
\end{figure}

\begin{figure}[!htp]
    \centering
    \includegraphics[scale=0.275]{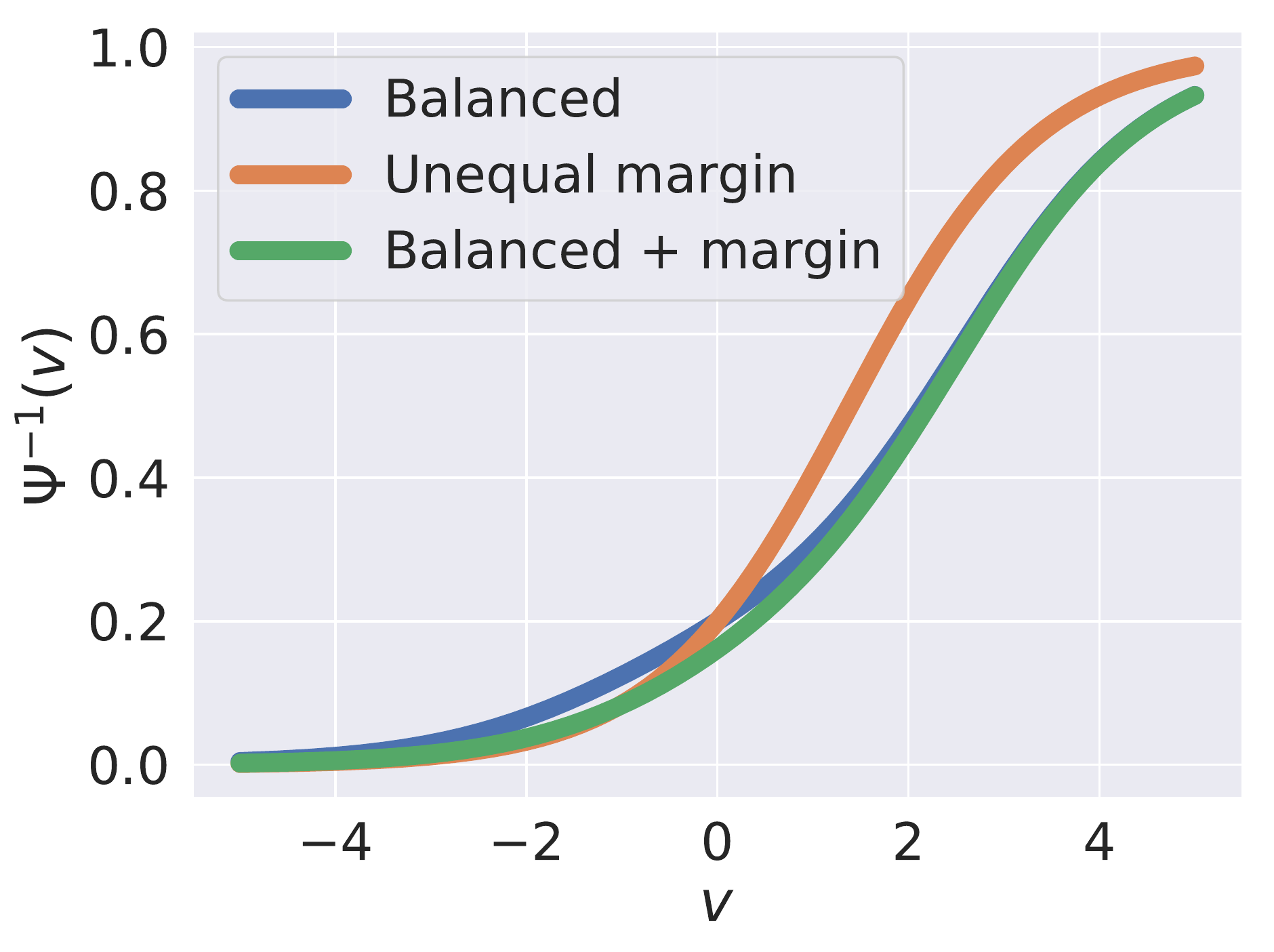}
    \includegraphics[scale=0.275]{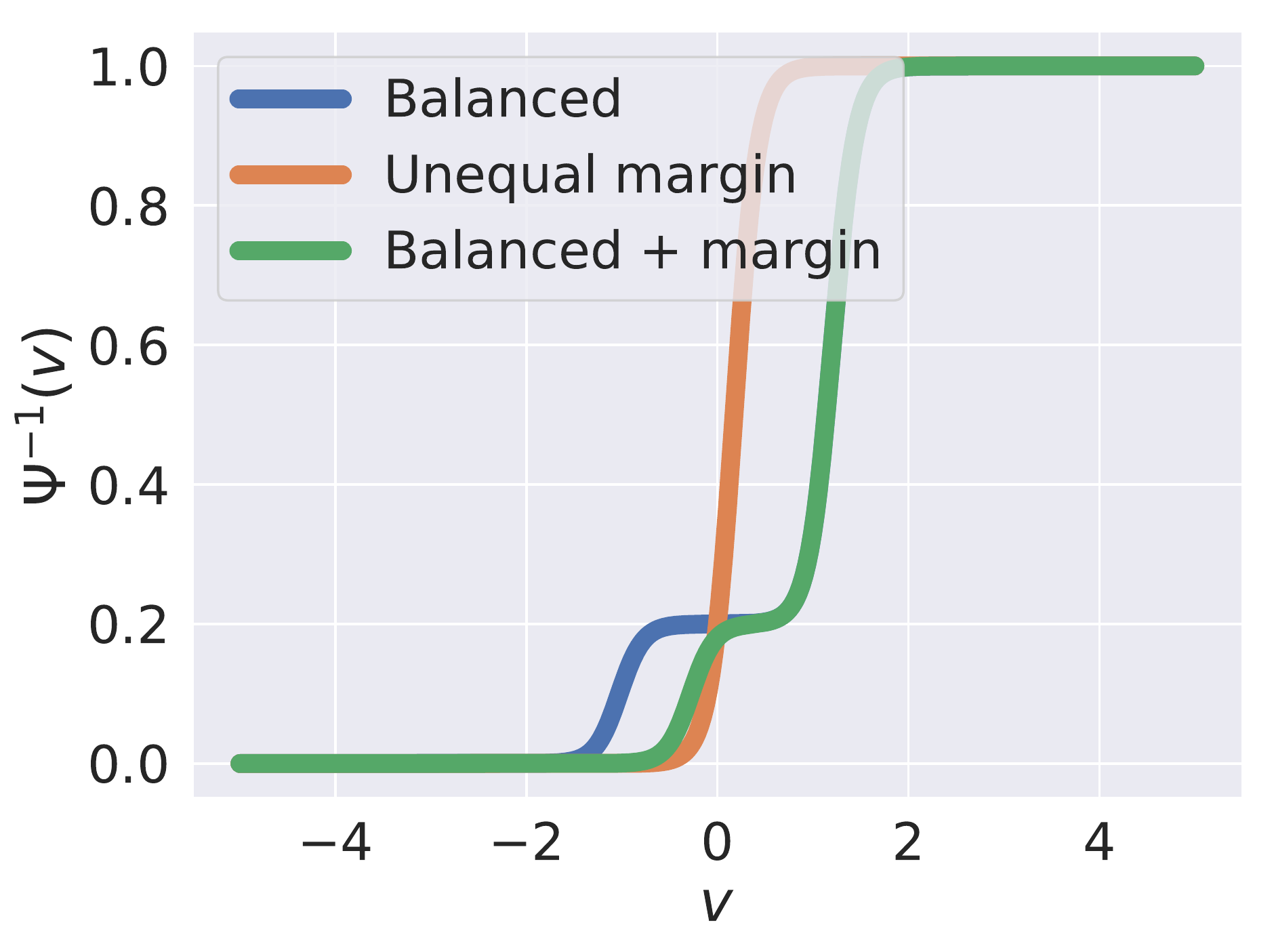}
    \caption{Comparison of link functions for various losses
    assuming $\pi = 0.2$,
    with $\gamma = 1$ (left) and $\gamma = 8$ (right).
    The balanced loss uses 
    $\omega_{y} = \frac{1}{\pi_{y}}$.
    The unequal margin loss uses 
    $\delta_{y} = \frac{1}{\gamma} \cdot \log \frac{1 - \pi_{y}}{\pi_{y}}$.
    The balanced + margin loss uses 
    $\delta_{-1} = \frac{\pi}{1-\pi}$, 
    $\delta_{+1} = 1$,
    $\omega_{+1} = \frac{1}{\pi}$.}
    \label{fig:inv_link_comparison}
\end{figure}

Figure~\ref{fig:link_comparison} and~\ref{fig:inv_link_comparison} compares the link functions for a few different settings:
\begin{itemize}[itemsep=0pt,topsep=0pt,leftmargin=16pt]
    \item the balanced loss, where 
    $\omega_{+1} = \frac{1}{\pi}$,
    $\omega_{-1} = \frac{1}{1 - \pi}$,
    and $\delta_{\pm 1} = 1$
    
    \item  an unequal margin loss, where 
    $\omega_{\pm 1} = 1$,
    $\delta_{+1} = \frac{1}{\gamma} \cdot \log \frac{1 - \pi}{\pi}$,
    and
    $\delta_{-1} = \frac{1}{\gamma} \cdot \log \frac{\pi}{1 - \pi}$ 
    

    \item a balanced + margin loss, where
    $\omega_{+1} = \frac{1}{\pi}$,
    $\omega_{-1} = \frac{1}{1 - \pi}$,    
    $\delta_{+1} = 1$,
    and
    $\delta_{-1} = \frac{\pi}{1-\pi}$.
\end{itemize}
The property 
{$\Psi^{-1}( 0 ) = \pi$ for $\pi = \Pr( y = 1 )$}
holds for the first two choices with any $\gamma > 0$,
and the third choice as $\gamma \to +\infty$.
This indicates the Fisher consistency of these losses for the balanced error.
However, the precise way this is achieved is strikingly different in each case.
In particular, each loss implicitly involves a fundamentally different link function.

\begin{figure}[!htp]
    \centering
    \includegraphics[scale=0.275]{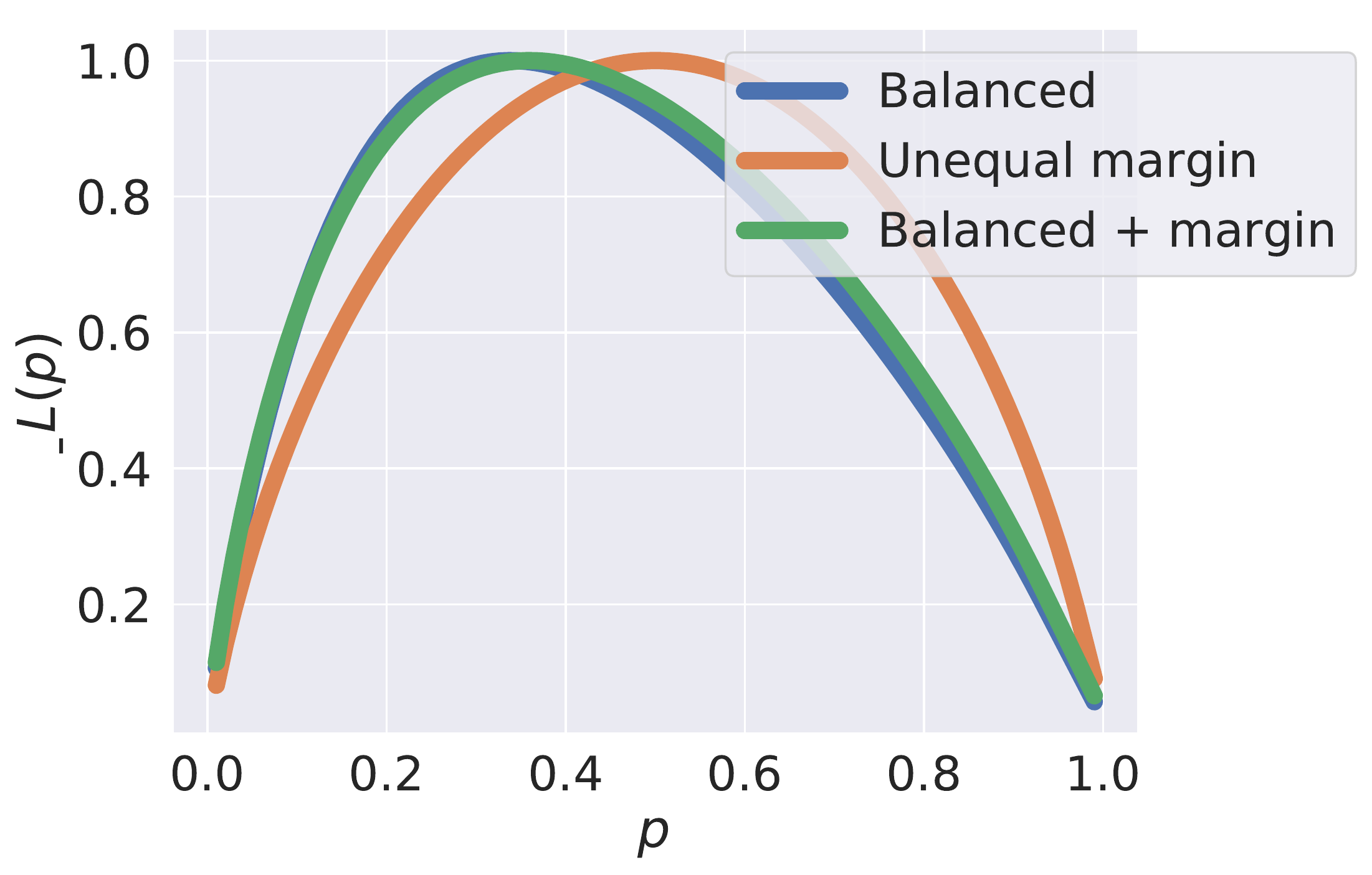}
    \includegraphics[scale=0.275]{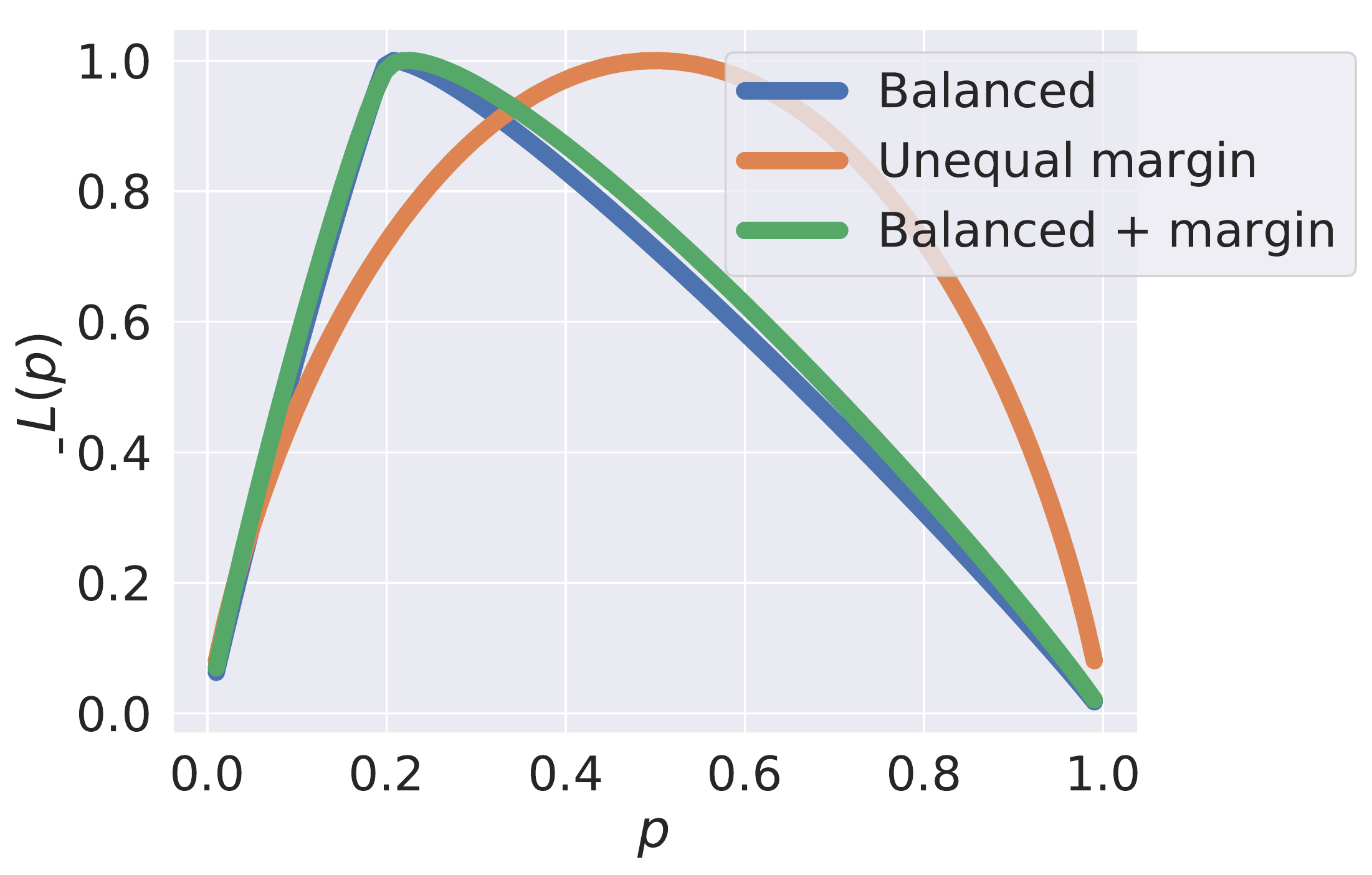}
    \caption{Comparison of conditional Bayes risk functions for various losses
    assuming $\pi = 0.2$,
    with $\gamma = 1$ (left) and $\gamma = 8$ (right).
    The balanced loss uses 
    $\omega_{y} = \frac{1}{\pi_{y}}$.
    The unequal margin loss uses 
    $\delta_{y} = \frac{1}{\gamma} \cdot \log \frac{1 - \pi_{y}}{\pi_{y}}$.
    The first balanced + margin loss uses $\delta_{-1} = \pi$, 
    $\delta_{+1} = 1$, $\omega_{+1} = \frac{1}{\pi}$.
    The second balanced + margin loss uses 
    $\delta_{-1} = \frac{\pi}{1-\pi}$, 
    $\delta_{+1} = 1$,
    $\omega_{+1} = \frac{1}{\pi}$.}
    \label{fig:bayes_risk_comparison}
\end{figure}

To better understand the effect of parameter choices,
Figure~\ref{fig:bayes_risk_comparison} illustrates the conditional Bayes risk curves,
i.e.,
$$ \underbar{L}( p ) = p \cdot \ell( +1, \Psi( p ) ) + (1 - p) \cdot \ell( +1, \Psi( p ) ). $$
We remark here that for the balanced error, this function takes the form $\underbar{L}( p ) = p \cdot \indicator{ p < \pi } + (1 - p) \cdot \indicator{ p > \pi }$,
i.e.,
it is a ``tent shaped'' concave function with a maximum at $p = \pi$.

For ease of comparison, we normalise this curves to have a maximum of $1$.
Figure~\ref{fig:bayes_risk_comparison} shows that simply applying unequal margins does \emph{not} affect the underlying conditional Bayes risk compared to the standard log-loss;
thus, the change here is purely in terms of the link function.
By contrast, either balancing the loss or applying a combination of weighting and margin modification results in a closer approximation to the conditional Bayes risk curve for the cost-sensitive loss with cost $\pi$.

%
\arxiv{}{%
\section{Relation to cost-sensitive SVMs}
\label{app:cs-svm}
We recapitulate the analysis of~\citet{Masnadi-Shirazi:2010} in our notation.
Consider a binary cost-sensitive learning problem with cost parameter $c \in ( 0, 1 )$.
The Bayes-optimal classifier for this task corresponds to $f^*( x ) = \indicator{ \eta( x ) > c }$.
The case $c = 0.5$ is the standard classification problem.

Suppose we wish to design a weighted, variable margin SVM for this task, i.e.,
\begin{align*}
    \ell( +1, f ) &= {\omega_{+1}} \cdot [ \delta_{+1} - f ]_+ \\
    \ell( -1, f ) &= {\omega_{-1}} \cdot [ \delta_{-1} + f ]_+
\end{align*}
where $\omega_{\pm 1}, \delta_{\pm 1} \geq 0$.
The conditional risk for this loss is
\begin{align*}
    L( \eta, f ) &= \eta \cdot \ell( +1, f ) + (1 - \eta) \cdot \ell( -1, f ) \\
    &= 
    \begin{cases}
    (1 - \eta) \cdot {\omega_{-1}} \cdot ( \delta_{-1} + f ) & \text{ if } f > \delta_{+1} \\
    \eta \cdot {\omega_{+1}} \cdot ( \delta_{+1} - f ) + (1 - \eta) \cdot {\omega_{-1}} \cdot ( \delta_{-1} + f ) & \text{ if } f \in [ -\delta_{-1}, \delta_{+1} ] \\
    \eta \cdot {\omega_{+1}} \cdot ( \delta_{+1} - f ) & \text{ if } f < -\delta_{-1}.
    \end{cases}
\end{align*}
As this is a piecewise linear function, 
which is decreasing for $f < -\delta_{-1}$ and
increasing for $f > \delta_{+1}$,
the only possible minimum is at $\{ \delta_{+1}, -\delta_{-1} \}$.
To ensure consistency, we seek the minimum to be $\delta_{+1}$ iff $\eta > c$.
Observe that
\begin{align*}
    L( \eta, \delta_{+1} ) < L( \eta, -\delta_{-1} ) &\iff (1 - \eta) \cdot {\omega_{-1}} < \eta \cdot \omega_{+1} \\
    &\iff \frac{\eta}{1 - \eta} > \frac{\omega_{-1}}{\omega_{+1}} \\
    &\iff \eta > \frac{\omega_{-1}}{\omega_{-1} + \omega_{+1}}.
\end{align*}
Consequently, we must have
$$ \frac{\omega_{-1}}{\omega_{-1} + \omega_{+1}} = c \iff \frac{\omega_{+1}}{\omega_{-1}} = \frac{1 - c}{c}. $$
Observe here that the margin terms $\delta_{\pm 1}$ do \emph{not} appear in the consistency condition:
thus, as long as the weights are suitably chosen, \emph{any} choice of margin terms will result in a consistent loss.

However,
the margins \emph{do} influence the form conditional Bayes risk:
this is
$$ \underbar{L}( \eta ) =
\begin{cases}
(1 - \eta) \cdot \omega_{-1} \cdot ( \delta_{-1} + \delta_{+1} ) & \text{ if } \eta > c \\
\eta \cdot \omega_{-1} \cdot ( \delta_{-1} + \delta_{+1} ) & \text{ if } \eta < c.
\end{cases}
$$
For the purposes of normalisation, it is natural to require this function to attain a maximum at $1$.
This corresponds to choosing
$$ \delta_{-1} + \delta_{+1} = \frac{1}{c} \cdot \frac{1}{\omega_{+1}}. $$

In the class-imbalance setting, $c = \pi$, and so we require
\begin{align*}
    \frac{\omega_{+1}}{\omega_{-1}} &= \frac{1 - \pi}{\pi} \\
    \delta_{-1} + \delta_{+1} &= \frac{1}{\pi} \cdot \frac{1}{\omega_{+1}}
\end{align*}
for consistency and normalisation respectively.
This gives two degrees of freedom: the choice of $\omega_{+1}$ (which determines $\omega_{-1}$), and
then the choice of $\delta_{+1}$ (which determines $\delta_{-1}$).
For example, we could pick 
$\omega_{+1} = \frac{1}{\pi}$, $\omega_{-1} = \frac{1}{1 - \pi}$,
$\delta_{+1} = 1$,
$\delta_{-1} = \frac{\pi}{1 - \pi}$.

To relate this to~\citet{Masnadi-Shirazi:2010}, the latter 
considered
separate costs $C_{-1}, C_{+1}$ for a false positive and false negative respectively.
With this, they suggested to use~\citet[Equation 34]{Masnadi-Shirazi:2010}
\begin{align*}
    \ell( +1, f ) &= d \cdot \left[ \frac{e}{d} - f \right]_+ \\
    \ell( -1, f ) &= a \cdot \left[ \frac{b}{a} + f \right]_+
\end{align*}
with 
$\delta_{+1} = \frac{e}{d} = 1$,
$d = \omega_{+1} = C_{+1}$,
$a = \omega_{-1} = 2 C_{-1} - 1$,
and $\delta_{-1} = \frac{b}{a} = \frac{1}{a}$.
The constraints $C_1 \geq 2 C_{-1} - 1$ and $C_{-1} \geq 1$ are also enforced.

Under this setup, the cost ratio is
$\frac{C_{-1}}{C_{-1} + C_{+1}}$.
In the class-imbalance setting, 
we have
$\frac{C_{-1}}{C_{-1} + C_{+1}} = \pi$, and so 
$C_{+1} = \frac{1-\pi}{\pi} \cdot C_{-1}$.
By the consistency condition, we have $C_{+1} = \omega_{+1} = \frac{1-\pi}{\pi} \cdot \omega_{-1} = \frac{1-\pi}{\pi} \cdot (2 C_{-1} - 1)$.
Thus, we must set $C_{-1} = 1$,
and so $C_{+1} = \frac{1 - \pi}{\pi}$.
Thus,
we obtain the parameters
$\omega_{+1} = \frac{1-\pi}{\pi}$,
$\omega_{-1} = 1$,
$\delta_{+1} = 1$,
$\delta_{-1} = \frac{\pi}{1 - \pi}$.
By rescaling the weights, we obtain
$\omega_{+1} = \frac{1}{\pi}$,
$\omega_{-1} = \frac{1}{1 - \pi}$,
$\delta_{+1} = 1$,
$\delta_{-1} = \frac{\pi}{1 - \pi}$.
Observe that this is exactly one of the losses considered in Appendix~\ref{sec:binary-consistency}.
}


%
\section{Experimental setup}
\label{app:architectures}
Intending a fair comparison, we use the same setup for all the methods for each dataset. 
All networks are trained with SGD with a momentum value of 0.9.
Unless otherwise specified,
linear learning rate warm-up is used in the first 5 epochs to reach the base learning rate,
and
a weight decay of $10^{-4}$ is used. 
Other dataset specific details are given below.

\textbf{CIFAR-10 and CIFAR-100}: 
We use a CIFAR ResNet-32 model trained for 200 epochs. The base learning rate is set to 0.1, which is decayed by 0.1 at the 160th epoch and again at the 180th epoch. Mini-batches of 128 images are used.

We also use the standard CIFAR data augmentation procedure used in previous works such as~\citet{Cao:2019, He:2016}, where 4 pixels are padded on each size and a random $32 \times 32$ crop is taken. Images are horizontally flipped with a probability of 0.5.

\textbf{ImageNet}: 
We use a ResNet-50 model trained for 90 epochs. 
The base learning rate is 0.4,
with cosine learning rate decay.
We use a batch size of 512 and the standard data augmentation comprising of random cropping and flipping as described in~\citet{Goyal:2017}.
Following~\citet{Kang:2020}, we use a weight decay of $5 \times 10^{-4}$ on this dataset.

\textbf{iNaturalist}: 
We again use a ResNet-50 and train it for 90 epochs with a base learning rate of 
0.4 
and cosine learning rate decay. 
The data augmentation procedure is the same as the one used in ImageNet experiment above.
\arxiv{}{We use a batch size of $512$.}


%
\section{Additional experiments}
We present here additional experiments:
\begin{enumerate}[label=(\roman*),itemsep=0pt,topsep=0pt,leftmargin=16pt]
    \item we present results for CIFAR-10 and CIFAR-100 on the {\sc Step} profile~\citep{Cao:2019} with $\rho = 100$
    
    \item we further verfiy that weight norms may not correlate with class priors under Adam
    
    \item we include the results of post-hoc correction, and a breakdown of per-class errors, on ImageNet-LT
\end{enumerate}

\subsection{Results on CIFAR-LT with {\sc Step}-100 profile}
\label{app:additional_results}

\begin{figure}[!ht]
    \centering
    \includegraphics[scale=0.325]{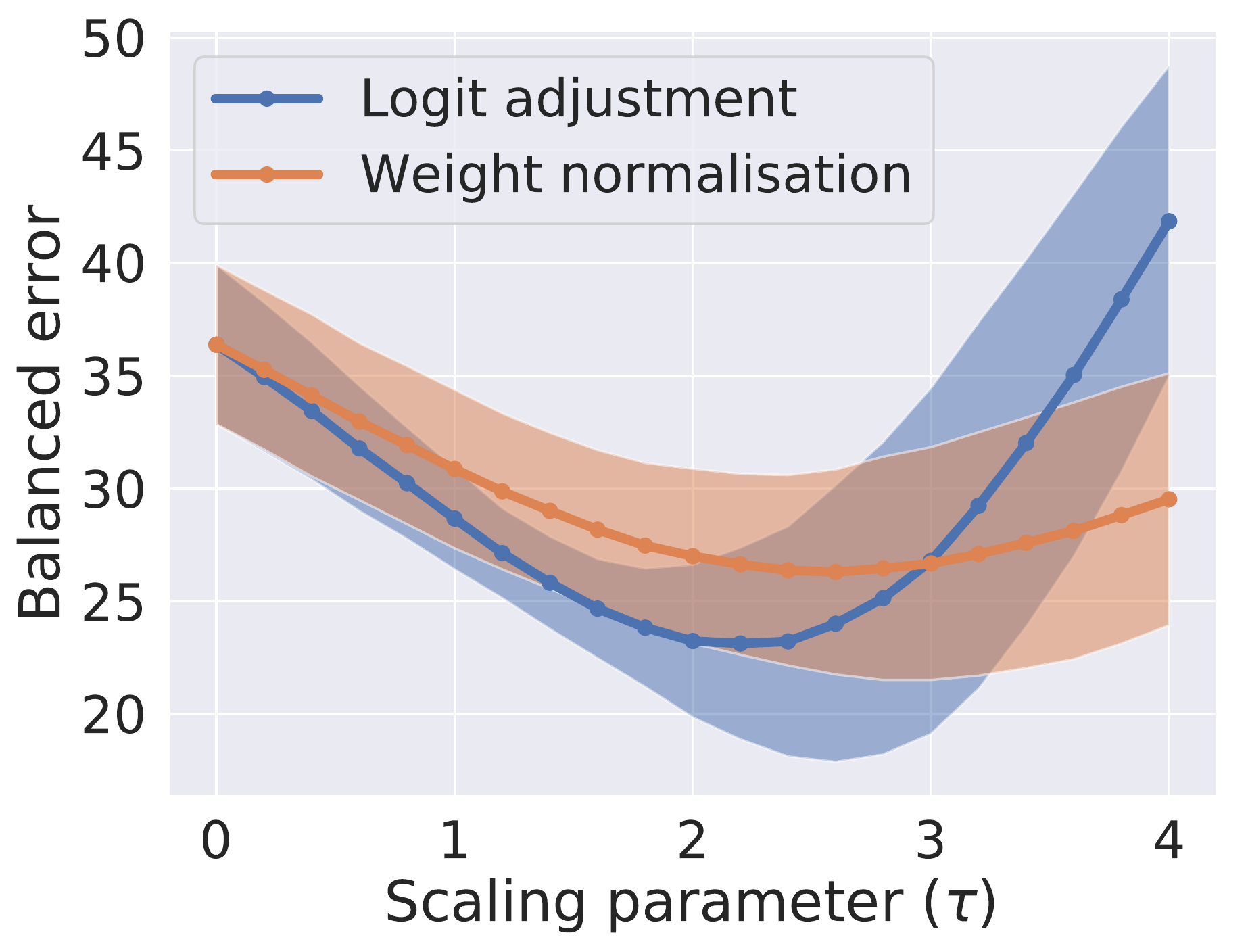}
    \includegraphics[scale=0.325]{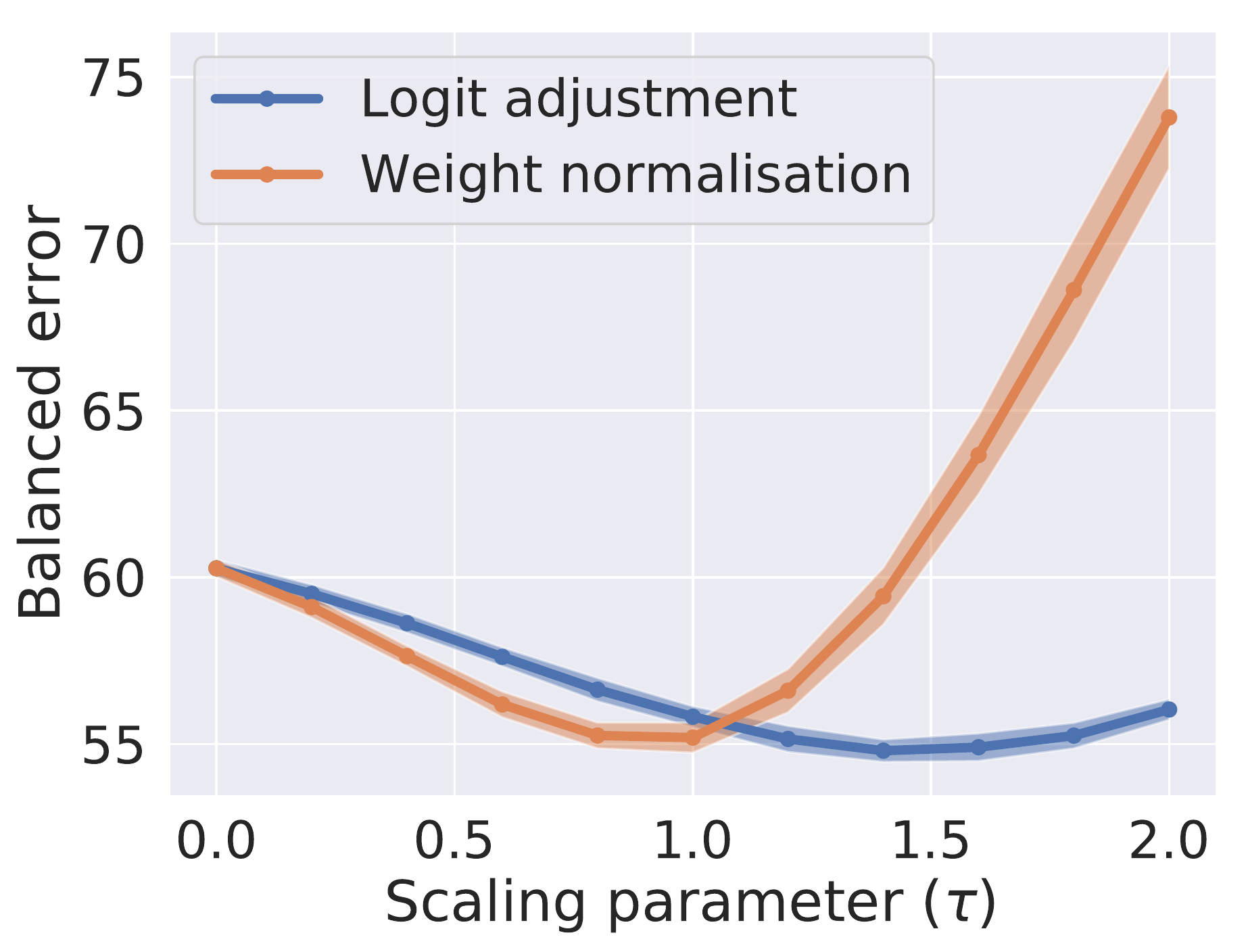}
    \caption{Post-hoc adjustment on {\sc Step}-100 profile, CIFAR-10 and CIFAR-100.
    Logit adjustment outperforms weight normalisation with suitable tuning.}
    \label{fig:posthoc_cifar100_step}
\end{figure}

Table~\ref{tbl:results_step} summarises results on the {\sc Step}-100 profile.
Here, with $\tau = 1$, weight normalisation slightly outperforms logit adjustment.
However, with $\tau > 1$, logit adjustment is again found to be superior (54.80);
see Figure~\ref{fig:posthoc_cifar100_step}.

\begin{table}[!ht]
    \centering
    \renewcommand{\arraystretch}{1.25}
    
    \begin{tabular}{@{}lll@{}}
        \toprule
        \textbf{Method} & \textbf{CIFAR-10-LT} & \textbf{CIFAR-100-LT} \\
        \toprule
        ERM                  & 
        36.54 &
        60.23 \\
        Weight normalisation ($\tau = 1$) & 
        30.86 &
        \best{55.19} \\
        Adaptive             & 
        34.61 &
        58.86 \\
        Equalised            & 
        31.42 &
        57.82 \\
        \midrule
        Logit adjustment post-hoc ($\tau = 1$) & 
        28.66 &
        55.82 \\
        Logit adjustment (loss)     & 
        \best{27.57} &
        55.52 \\
        \bottomrule
    \end{tabular}

    \caption{
    Test set balanced error (averaged over $5$ trials) on
    CIFAR-10-LT
    and
    CIFAR-100-LT under the {\sc Step}-100 profile;
    lower is better.
    On CIFAR-100-LT, weight normalisation edges out logit adjustment.
    See Figure~\ref{fig:posthoc_cifar100_step} for a demonstrated that tuned versions of the same outperfom weight normalisation.
    }
    \label{tbl:results_step}
\end{table}

%
\subsection{Per-class errors on ImageNet-LT}

\begin{figure}[!ht]
    \centering
    \includegraphics[scale=0.35]{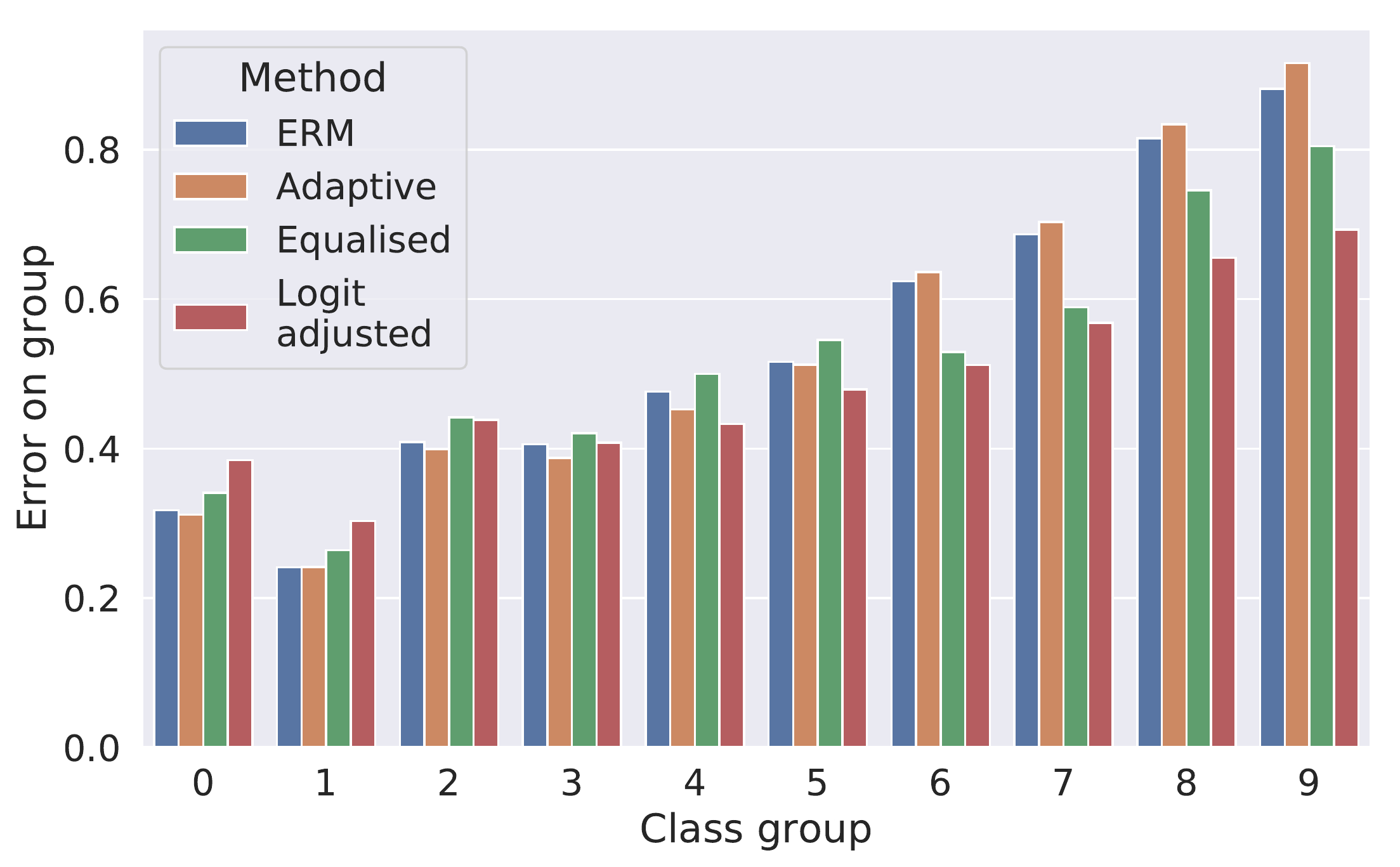}
    \caption{Comparison of per-class balanced error on ImageNet-LT.
    Classes are sorted in order of frequency, and bucketed into 10 groups.}
    \label{fig:per_class_imagenet}
\end{figure}

Figure~\ref{fig:per_class_imagenet} breaks down the per-class accuracies on ImageNet-LT.
As before, the logit adjustment procedure shows significant gains on rarer classes.

\subsection{Post-hoc correction on ImageNet-LT}

\begin{figure}[!ht]
    \centering
    \includegraphics[scale=0.5]{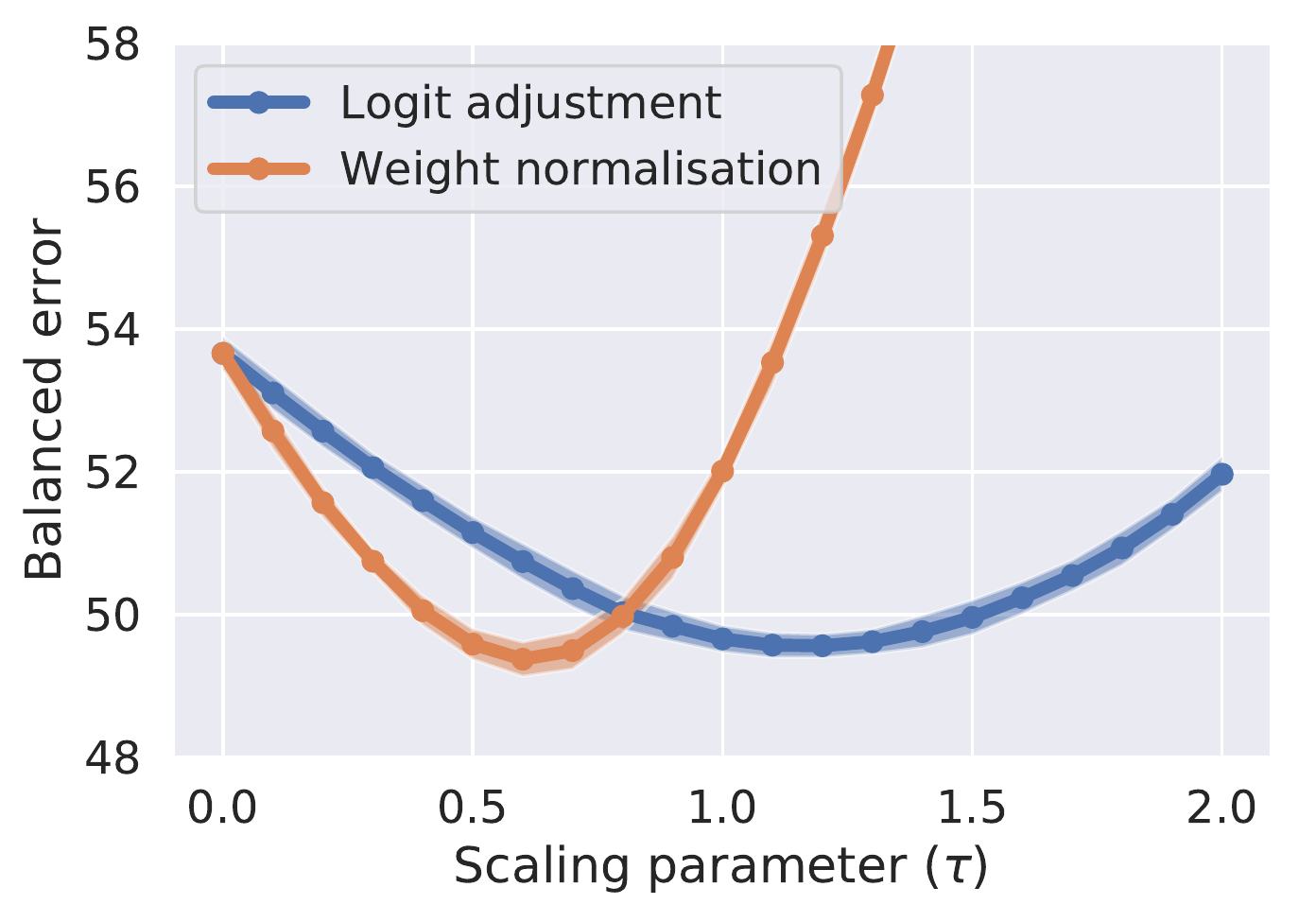}
    \caption{Post-hoc correction on ImageNet.}
    \label{fig:imgnet_lt_post_hoc_comparison}
\end{figure}

Figure~\ref{fig:imgnet_lt_post_hoc_comparison} compares post-hoc correction techniques as the scaling parameter $\tau$ is varied on ImageNet-LT.
As before, logit adjustment with suitable tuning is seen to be competitive with weight normalisation.

\subsection{Per-group errors}

\begin{figure}[!ht]
    \centering
    
    \subcaptionbox{CIFAR-10-LT}{\includegraphics[scale=0.25]{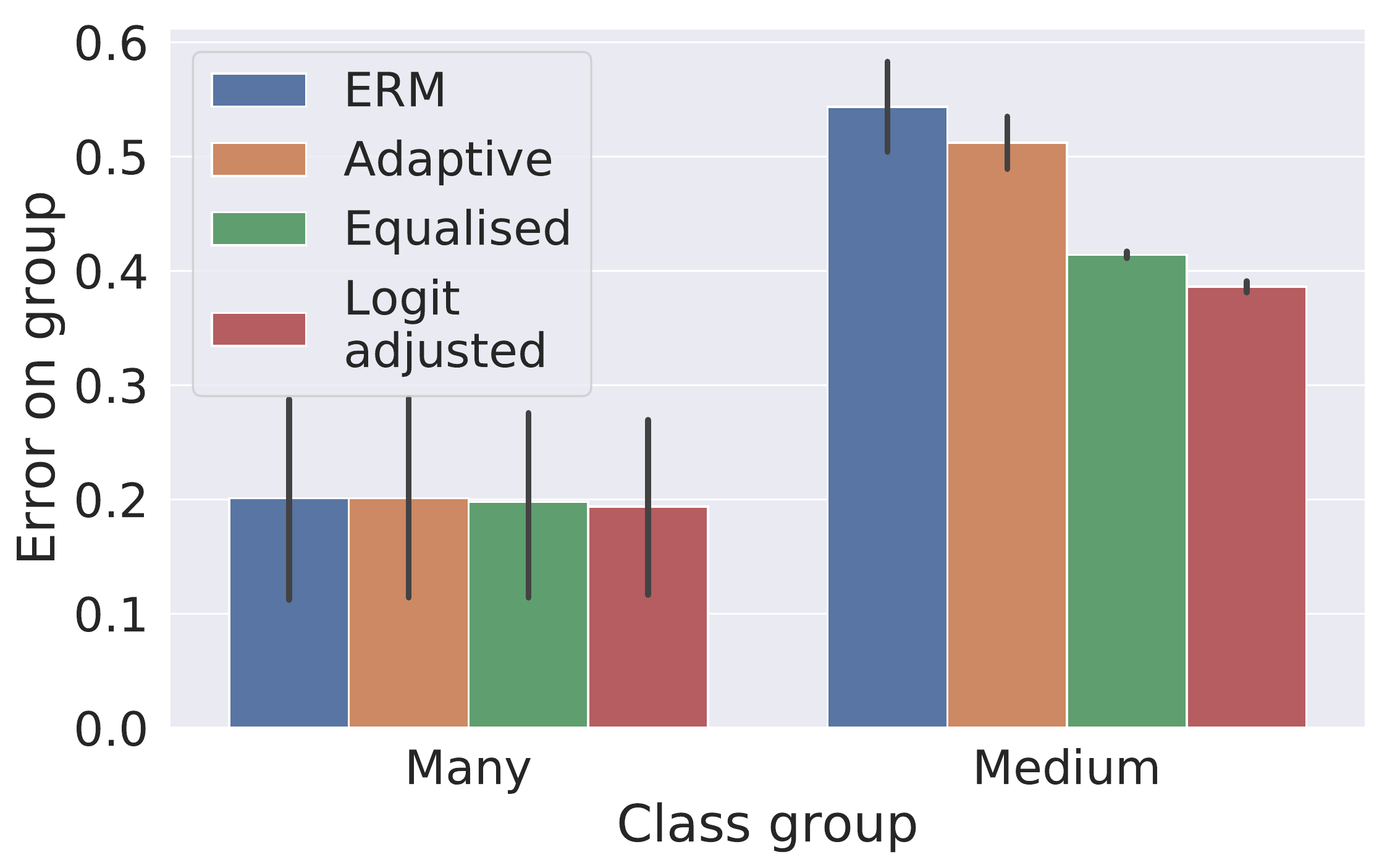}}
    \qquad
    \subcaptionbox{CIFAR-100-LT.}{\includegraphics[scale=0.25]{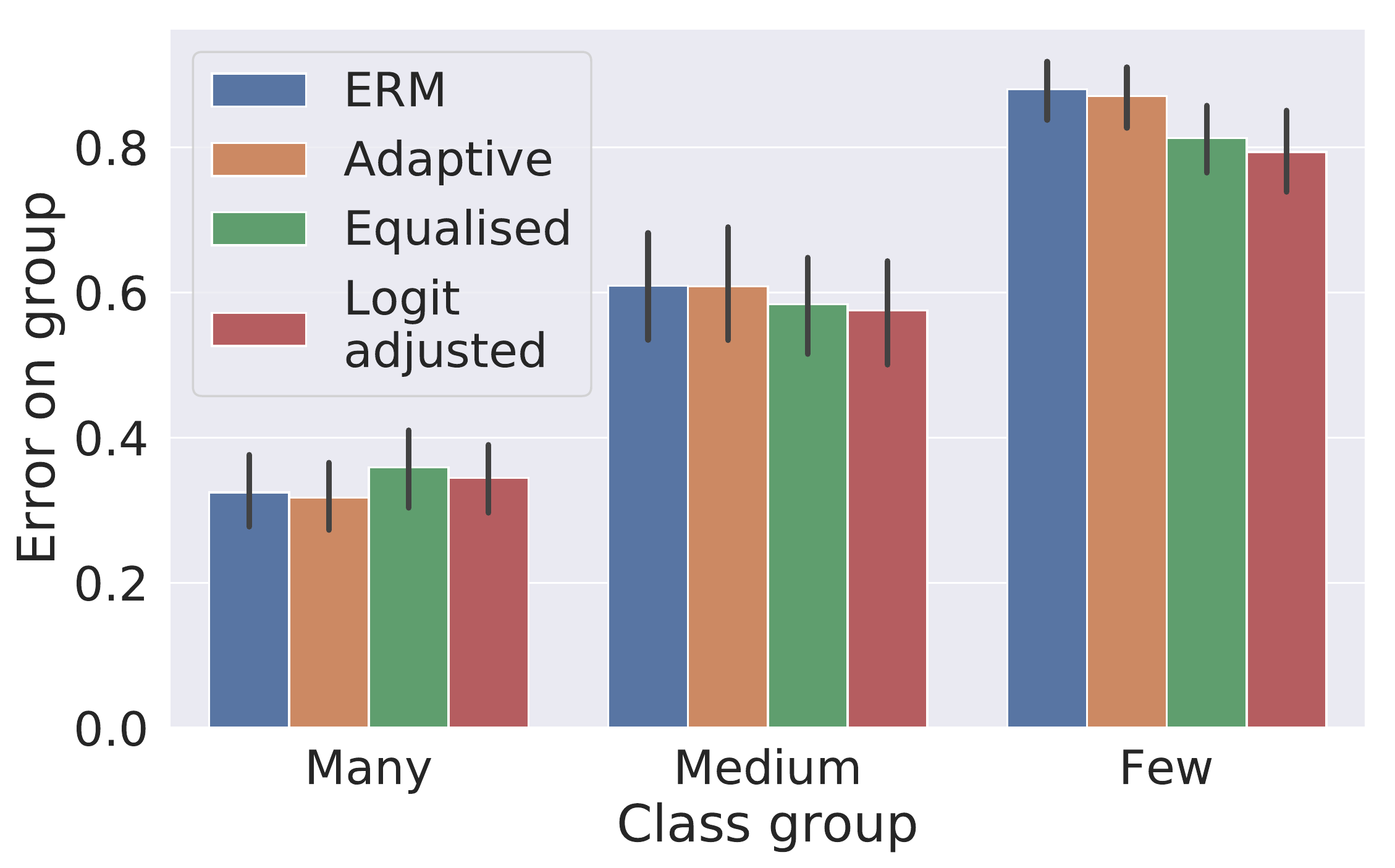}}
    
    \subcaptionbox{ImageNet-LT.}{\includegraphics[scale=0.25]{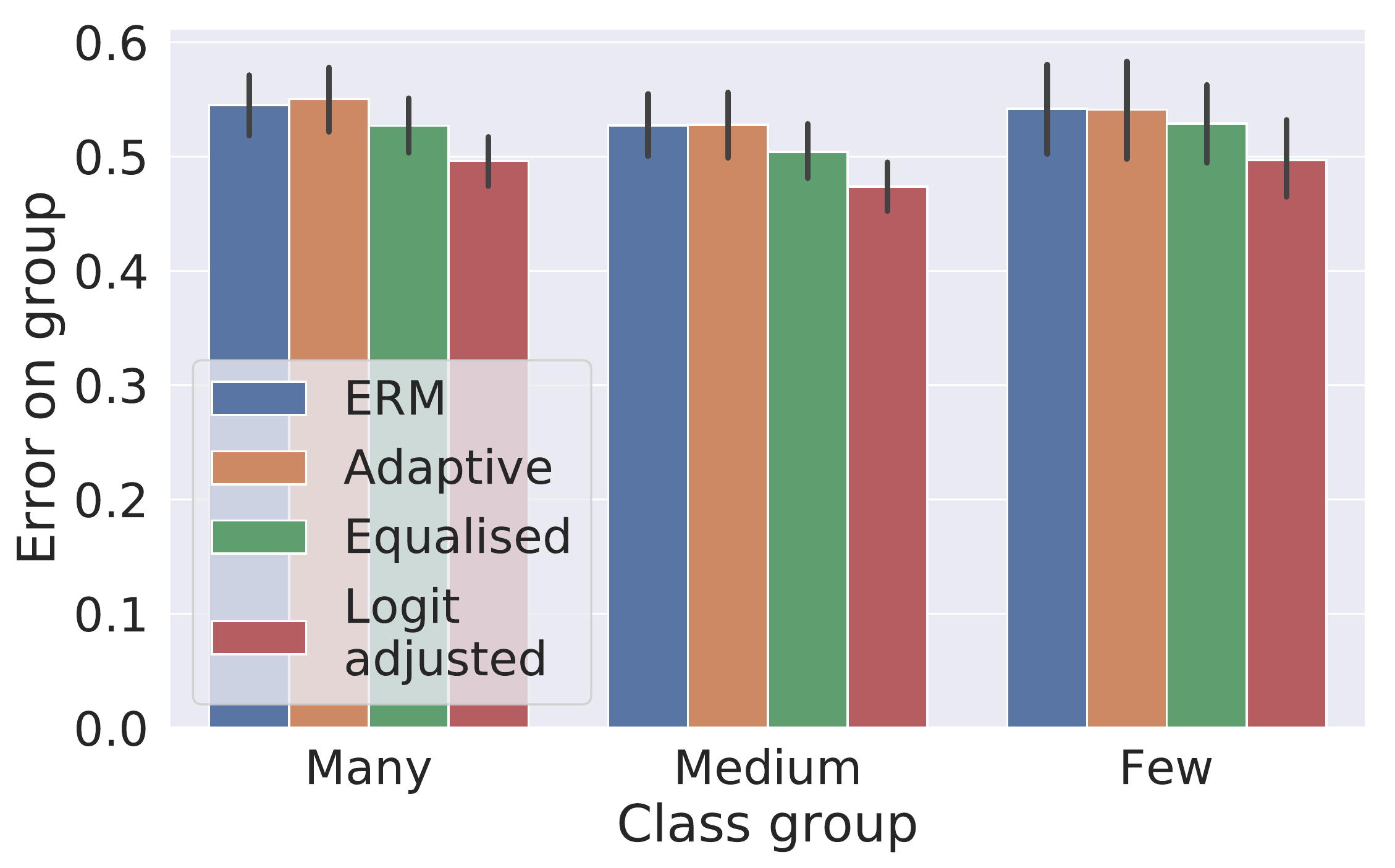}}
    \qquad
    \subcaptionbox{iNaturalist.}{\includegraphics[scale=0.25]{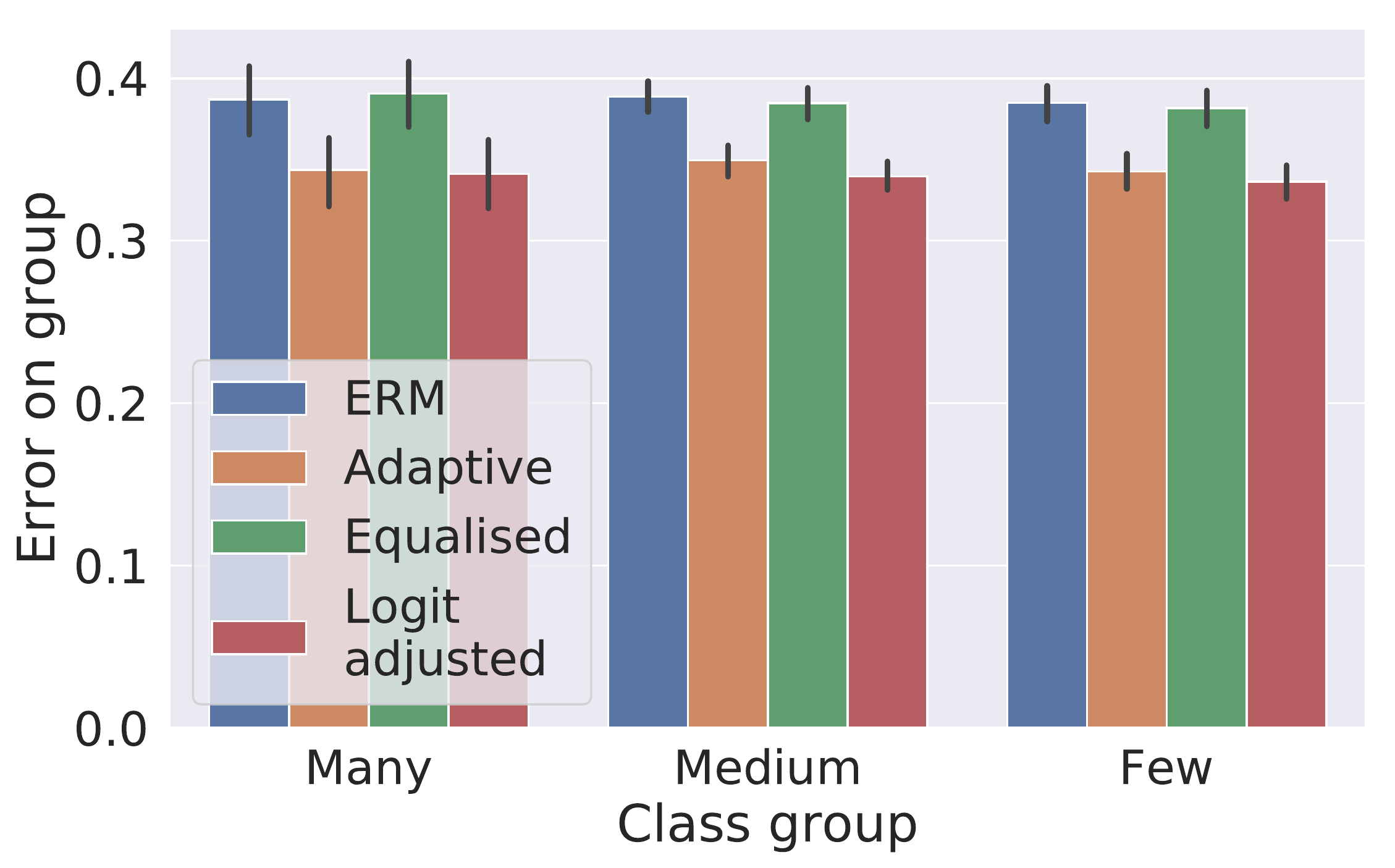}}
    
    \caption{Comparison of per-group errors.
    We construct three groups of classes:
``Many'', comprising those with at least 100 training examples;
``Medium'', comprising those with at least 20 and at most 100 training examples;
and
``Few'', comprising those with at most 20 training examples.}
    \label{fig:per_group_errors}
\end{figure}

Following~\citet{Liu:2019,Kang:2020}, we additionally report errors on a per-group basis,
where we construct three groups of classes:
``Many'', comprising those with at least 100 training examples;
``Medium'', comprising those with at least 20 and at most 100 training examples;
and
``Few'', comprising those with at most 20 training examples.
This is a coarser level of granularity than the grouping employed in the previous section, and the body.
Figure~\ref{fig:per_group_errors} shows that 
the logit adjustment procedure shows consistent gains over all three groups.

\label{sec:additional_experiments}


%
\arxiv{%
\section{Summary of existing work}

Table~\ref{tbl:literature} provides a summary of different strands of the class imbalance literature.
\begin{table}[!ht]
    \centering
    \begin{tabular}{@{}llp{2.75in}@{}}
        \toprule
        {\bf Family} & {\bf Method} & {\bf Reference} \\
        \toprule
        Post-hoc correction   & Modify threshold & \citep{Fawcett:1996,Provost:2000,Maloof03,King:2001,Collell:2016} \\
        & Normalise weights & \citep{Zhang:2019,Kim:2019,Kang:2020} \\
        \midrule
        Data modification & Under-sampling & \citep{KubatMa97,Wallace:2011} \\
        & Over-sampling & \citep{Chawla:2002} \\
        & Feature transfer & \citep{Yin:2018} \\
        \midrule
        Loss weighting & Loss balancing & \citep{Xie:1989,Morik:1999,Menon:2013} \\
        & Volume weighting & \citep{Cui:2019} \\ 
        & Average top-$k$ loss & \citep{Fan:2017} \\
        & Domain adaptation & \citep{Jamal:2020} \\
        \midrule
        Margin modification & Cost-sensitive SVM & \citep{Masnadi-Shirazi:2010,Iranmehr:2019} \\
        & Range loss & \citep{Zhang:2017} \\
        & Label-aware margin & \citep{Cao:2019}
        \\
        & Equalised negatives & \citep{Tan:2020} \\
         \bottomrule
    \end{tabular}
    \caption{Summary of different approaches to learning under class imbalance.}
    \label{tbl:literature}
\end{table}
}{}


%
\section{Does weight normalisation increase margins?}
\label{sec:normalisation-margins}

Suppose that one uses SGD with a momentum, and finds solutions where $\| w_y \|_2$ tracks the class priors.
One intuition behind normalisation of weights 
is that,
drawing inspiration from the binary case,
this ought to 
increase the classification margins for tail classes.

Unfortunately, this intuition is \emph{not} necessarily borne out.
Consider a scorer $f_y( x ) = w_{y}^{\mathrm{T}} \Phi( x )$, where $w_{y} \in \Real^{d}$ and $\Phi \colon \XCal \to \Real^d$.
The \emph{functional} margin for an example $(x, y)$ is~\citep{Koltchinskii:2000}
\begin{align}
\label{eq:margin_mc}
\gamma_f(x, y) \defEq w_y ^{\mathrm{T}} \Phi(x) - \max_{y' \neq y} w_{y'}^{\mathrm{T}} \Phi(x).
\end{align}
This generalises the classical binary margin,
wherein by convention $\YCal = \{ \pm 1 \}$, $w_{-1} = -w_1$,
and
\begin{align}
\label{eq:f_margin_bin}
\gamma_{\rm f}(x, y) \defEq y \cdot w_{1}^{\mathrm{T}} \Phi(x) = {\frac{1}{2}} \cdot \left( w_{y}^{\mathrm{T}} \Phi(x) - w_{-y}^{\mathrm{T}} \Phi(x)\right),
\end{align}
which agrees with \eqref{eq:margin_mc} upto scaling.
One may also define the \emph{geometric} margin in the binary case 
to be 
the distance of $(x, y)$ from its classifier:
\begin{align}
\gamma_{g, {\rm b}}( x ) \defEq \frac{|w_{1}\cdot\Phi(x)|}{\|w_{1}\|_2}.
\end{align}
Clearly, $\gamma_{g, {\rm b}}( x ) = \frac{| \gamma_{\rm f}(x, y) |}{ \| w_{1} \|_2 }$,
and so for fixed functional margin,
one may increase the geometric margin by minimising $\| w_1 \|_2$.
However,
the same is \emph{not} necessarily true in the multiclass setting,
since here the functional and geometric margins do not generally align~\citep{Tatsumi:2011,Tatsumi:2014}.
In particular, controlling each $\| w_y \|_2$ does \emph{not} necessarily control the geometric margin.


%
\section{Bayes-optimal classifier under Gaussian class-conditionals}
\label{app:gaussian_ber}

Suppose
$$ \Pr( x \mid y ) = \frac{1}{\sqrt{2 \pi} \sigma} \cdot \exp\left( -\frac{\| x - \mu_y \|^2}{2 \sigma^2} \right) $$
for suitable $\mu_y$ and $\sigma$.
Then,
\begin{align*}
    \Pr( x \mid y = +1 ) > \Pr( x \mid y = -1 ) &\iff \exp\left( -\frac{\| x - \mu_{+1} \|^2}{2 \sigma^2} \right) > \exp\left( -\frac{\| x - \mu_{-1} \|^2}{2 \sigma^2} \right) \\
    &\iff \frac{\| x - \mu_{+1} \|^2}{2 \sigma^2} < \frac{\| x - \mu_{-1} \|^2}{2 \sigma^2} \\
    &\iff {\| x - \mu_{+1} \|^2} < {\| x - \mu_{-1} \|^2} \\
    &\iff 2 \cdot ( \mu_{+1} - \mu_{-1} )^{\mathrm{T}} x > \| \mu_{+1} \|^2 -  \| \mu_{-1} \|^2.
\end{align*}
Now use the fact that in our setting, $\| \mu_{+1} \|^2 = \| \mu_{-1} \|^2$.

We remark also that the class-probability function is
\begin{align*}
    \Pr( y = +1 \mid x ) &= \frac{\Pr( x \mid y = +1 ) \cdot \Pr( y = +1 )}{\Pr( x )} \\
    &= \frac{\Pr( x \mid y = +1 ) \cdot \Pr( y = +1 )}{\sum_{y'} \Pr( x \mid y' ) \cdot \Pr( y' )} \\
    &= \frac{1}{1 + \frac{\Pr( x \mid y = -1 ) \cdot \Pr( y = -1 )}{\Pr( x \mid y = +1 ) \cdot \Pr( y = +1 )}}.
\end{align*}
Now,
\begin{align*}
    \frac{\Pr( x \mid y = -1 )}{\Pr( x \mid y = +1 )} &= \exp\left( \frac{\| x - \mu_{+1} \|^2 - \| x - \mu_{-1} \|^2}{2 \sigma^2} \right) \\
    &= \exp\left( \frac{ \| \mu_{+1} \|^2 -  \| \mu_{-1} \|^2 - 2 \cdot ( \mu_{+1} - \mu_{-1} )^{\mathrm{T}} x }{2 \sigma^2} \right) \\
    &= \exp\left( \frac{ - ( \mu_{+1} - \mu_{-1} )^{\mathrm{T}} x }{\sigma^2} \right).
\end{align*}
Thus,
$$ \Pr( y = +1 \mid x ) = \frac{1}{1 + \exp(-w_*^{\mathrm{T}} x + b_*)}, $$
where $w_* = \frac{1}{\sigma^2} \cdot ( \mu_{+1} - \mu_{-1} )$,
and $b_* = \log \frac{\Pr( y = -1 )}{\Pr( y = +1 )}$.
This implies that a sigmoid model for $\Pr( y = +1 \mid x )$, as employed by logistic regression, is well-specified for the problem.
Further, the bias term $b_*$ is seen to take the form of the log-odds of the class-priors per~\eqref{eqn:bayes-logit-adjustment}, as expected.


\end{document}